# Development of a Deep Learning Based Control Framework for Exoskeleton Robots

**SK Hasan**
hasansk@miamioh.edu
Department of Mechanical and Manufacturing Engineering
Miami University
Oxford, OH 45056

# Development of a Deep Learning Based Control Framework for Exoskeleton Robots

**Abstract:**

The purpose of this study is to develop a computationally efficient deep learning based control framework for high degree of freedom exoskeleton robots to address the real time computational limitations associated with conventional model based control. A parallel structured deep neural network was designed for a seven degree of freedom human lower extremity exoskeleton robot. The network consists of four layers with 49 densely connected neurons and was trained using physics based data generated from the analytical dynamic model. During real time implementation, the trained neural network predicts joint torque commands required for trajectory tracking, while a proportional derivative controller compensates for residual prediction errors. Stability of the proposed control scheme was analytically established, and robustness to parameter variations was evaluated using analysis of variance. Comparative simulations were conducted against computed torque, model reference computed torque, sliding mode, adaptive, and linear quadratic controllers under identical robot dynamics. Results demonstrate accurate trajectory tracking with torque profiles comparable to conventional nonlinear controllers while reducing computational burden. These findings suggest that the proposed deep learning based hybrid controller offers an efficient and robust alternative for controlling multi degree of freedom exoskeleton robots.

# Development of a Deep Learning-Driven Control Framework for Exoskeleton Robots

## 1. Introduction

Exoskeleton robots have attracted significant research interest due to their potential to augment human physical capabilities, advance rehabilitation technologies, and improve the quality of life for individuals with mobility impairments [1-3]. As wearable robotic systems, exoskeletons assist and enhance human motion, enabling activities such as walking, load carrying, and balance maintenance. Their demonstrated effectiveness in human augmentation and rehabilitation has led to a rapid increase in research efforts in this field [4-6].

Model-based control remains a dominant approach in robotic control applications, as it explicitly incorporates the robot's kinematic and dynamic models into the control algorithm. Robot dynamic equations of motion are commonly derived using either the Newton–Euler formulation or the Lagrangian energy method [7]. Although these approaches differ in formulation, they yield equivalent dynamic representations. Robot dynamics are inherently nonlinear due to configuration-dependent mass and inertia, gravitational effects, Coriolis and centrifugal forces, and joint friction. A structured dynamic model enables these nonlinear components to be explicitly identified and compensated.

In model-based control, nonlinear terms such as gravity, Coriolis–centrifugal effects, and joint friction are computed and fed back to the system to cancel their influence, while the control input is scaled by the configuration-dependent mass matrix to compensate for inertia-related nonlinearities. Following this linearization, joint accelerations become proportional to the applied torques, with the mass matrix acting as the proportionality factor. Consequently, conventional linear control schemes can be applied to regulate robot motion. This framework represents a canonical example of model-based control.

A wide range of robot control strategies has been reported in the literature, including linear [8, 9], nonlinear [10], robust [11], optimal [9], and adaptive control schemes [12]. A common characteristic of most established approaches is their reliance on accurate dynamic models for both linearization and control. However, as the number of degrees of freedom (DOFs) increases, robot dynamics become increasingly complex and computationally demanding. Many advanced control strategies operate in the joint torque domain, which requires significantly higher sampling rates than position- or velocity-based control methods [8, 9, 11-15]. Executing the full dynamic model within a strict sampling interval therefore demands substantial computational resources.

Moreover, the mass, gravity, and Coriolis–centrifugal matrices in the robot's dynamic equations are sequentially structured, limiting the effectiveness of multicore CPUs and GPUs in reducing execution time. As a result, increasing computational speed through parallel hardware alone is insufficient. In contrast, a neural network-based equivalent dynamic model can provide a computationally efficient, parallel-structured approximation, enabling the combined use of CPU and GPU architectures to accelerate control algorithm execution.

Deep learning has emerged as a powerful tool for classification, regression, and prediction problems across a wide range of domains [16]. A comprehensive overview of deep neural network applications is provided in reference 17 [17]. In robotics, neural networks have been extensively explored for dynamic modeling and control. For example, Chen *et al.* [18] integrated iterative learning control with a neural network to improve trajectory tracking in multi-DOF industrial robots, using physical excitation to identify internal dynamics and a neural network-based inverse model for feedforward compensation. Su *et al.* [19] employed adaptive neural networks to compensate for nonlinearities and friction in a single-link flexible manipulator, followed by linear control. Related work on neural network-assisted control of flexible manipulators is reported in [20-25].

Lin *et al.* [26] developed a deep neural network for gravity and disturbance compensation, employing knowledge distillation to enable real-time implementation. Mukhopadhyay *et al.* [27] investigated recurrent neural networks for robot dynamic modeling using sensor data and compared SRNN, LSTM, and GRU architectures. Similarly, Liu *et al.* [28] used LSTM networks to learn inverse robot dynamics based on analytically generated torque data. Liang *et al.* [29] applied deep learning to online robot dynamic modeling using real-time physical excitation data. Panwar *et al.* [30] utilized neural networks for generating human-like bipedal robot trajectories, while Narendra and Parthasarathy [31] demonstrated the effectiveness of recurrent neural networks for identification and control of nonlinear dynamical systems. Polydoros *et al.* [32] proposed a real-time deep learning framework for robotics capable of learning from noisy data with fast convergence. Neural networks have also been widely adopted in adaptive robot control, as evidenced by numerous studies [33-37].

Despite these advances, a systematic framework for generating physics-governed training data for inverse dynamic modeling of multi-DOF robotic systems, combined with a deep neural network–based dynamic controller, remains insufficiently explored. This gap motivates the contributions of this work.

The main contributions of this article are as follows:

1. A physics-based methodology for generating training data suitable for machine learning-based control of multi-DOF robotic systems.

2. The development of a computationally efficient deep neural network-based hybrid controller for serial-link robots that enables parallel CPU-GPU computation and achieves high trajectory tracking performance comparable to conventional nonlinear control techniques.

The remainder of this article is organized as follows. Section 1 reviews the state of the art in robot control with emphasis on neural network–based methods, formulates the problem, and outlines the proposed solution. Section 2 presents the dynamic modeling of the human lower-extremity exoskeleton robot. Section 3 describes the computed torque control approach. Section 4 details the dynamic simulation, data generation, neural network architecture, training, and validation. Section 5 introduces the proposed hybrid controller and presents its stability analysis. Section 6 discusses simulation results. Section 7 evaluates robustness to parameter variations using analysis of variance (ANOVA). Section 8 presents comparative studies with computed torque control, model reference computed torque control, sliding mode control, adaptive control, and linear quadratic regulation under identical robot dynamics.

## 2. Dynamic Modeling of the Exoskeleton Robot

The modified Denavit-Hartenberg (DH) convention  was used to assign coordinate frames to each degree of freedom (Figure 1), and the corresponding DH parameter table was formulated (Table 1). Substituting the DH parameters into the general homogeneous transformation matrix given in (Equation. (1)) yields the individual link transformation matrices, as expressed in equations (2)-(8). The complete forward kinematic model was obtained by sequentially multiplying the seven link transformation matrices, as shown in Equation (9). The resulting forward kinematics defines the position and orientation of the end-effector frame with respect to the base frame.

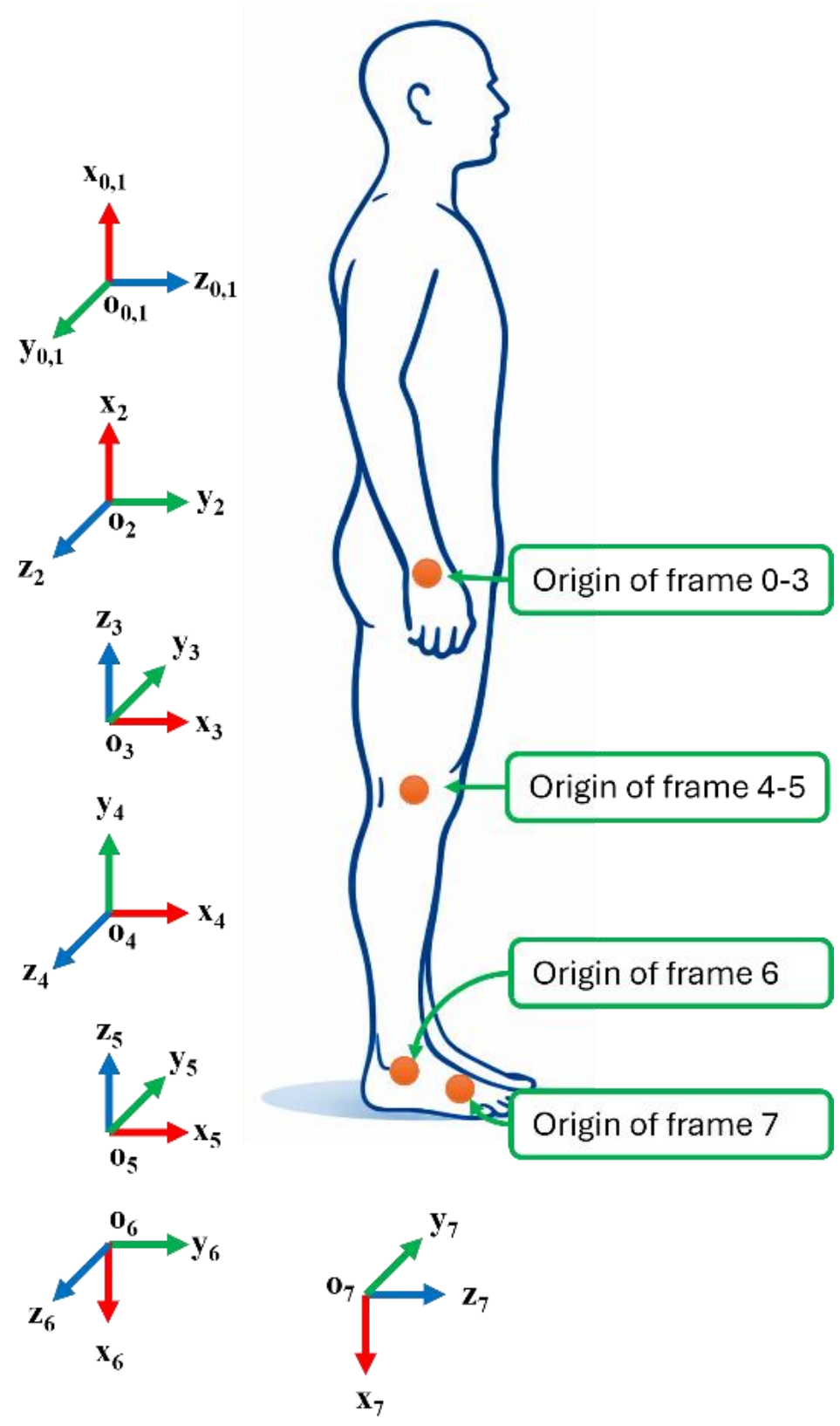

Figure 1 Link Frame assignment based on modified D-H parameter

Table 1: Human lower extremity DH parameters

| Joint (i) | | Joint name | Joint variable | Link offset | Link length | Link twist |
|---|---|---|---|---|---|---|
| | | | $\boldsymbol{\theta_i}$ | $\boldsymbol{d_i}$ | $\boldsymbol{a_{i-1}}$ | $\boldsymbol{\alpha_{i-1}}$ |
| 1. | | Abduction/Adduction | $\theta_1$ | 0 | 0 | 0 |
| 2. | Hip | Flexion/Extension | $\theta_2 - \frac{\pi}{2}$ | 0 | 0 | $-\frac{\pi}{2}$ |
| 3. | | Internal/External rotation | $\theta_3$ | $-l_1$ | 0 | $-\frac{\pi}{2}$ |
| 4. | Knee | Flexion/Extension | $\theta_4$ | 0 | 0 | $\frac{\pi}{2}$ |
| 5. | | Internal rotation | $\theta_5$ | $-l_2$ | 0 | $-\frac{\pi}{2}$ |
| 6. | Ankle | Dorsiflexion/Plantarflexion | $\theta_6 - \frac{\pi}{2}$ | 0 | 0 | $\frac{\pi}{2}$ |
| 7. | | Pronation/Supination | $\theta_7$ | 0 | $a_1$ | $-\frac{\pi}{2}$ |

$$ {}^{i-1}_{\;\,i}T = \begin{bmatrix} \cos\theta_i & -\sin\theta_i & 0 & \alpha_{i-1} \\ \sin\theta_i \cos\alpha_{i-1} & \cos\theta_i \cos\alpha_{i-1} & -\sin\alpha_{i-1} & -\sin\alpha_{i-1}\, d_i \\ \sin\theta_i \sin\alpha_{i-1} & \cos\theta_i \sin\alpha_{i-1} & \cos\alpha_{i-1} & \cos\alpha_{i-1} d_i \\ 0 & 0 & 0 & 1 \end{bmatrix} \tag{1} $$

$$ {}^{0}T_1 = \begin{bmatrix} \cos(\theta_1) & -\sin(\theta_1) & 0 & 0 \\ \sin(\theta_1) & \cos(\theta_1) & 0 & 0 \\ 0 & 0 & 1 & 0 \\ 0 & 0 & 0 & 1 \end{bmatrix} \tag{2} $$

$$ {}^{1}T_2 = \begin{bmatrix} \sin(\theta_2) & \cos(\theta_2) & 0 & 0 \\ 0 & 0 & 0 & 0 \\ \cos(\theta_2) & -\sin(\theta_2) & 1 & 0 \\ 0 & 0 & 0 & 1 \end{bmatrix} \tag{3} $$

$$ {}^{2}T_3 = \begin{bmatrix} \cos(\theta_3) & -\sin(\theta_3) & 0 & 0 \\ 0 & 0 & 0 & -l_1 \\ -\sin(\theta_3) & -\cos(\theta_3) & 1 & 0 \\ 0 & 0 & 0 & 1 \end{bmatrix} \tag{4} $$

$$ {}^{3}T_4 = \begin{bmatrix} \cos(\theta_4) & -\sin(\theta_4) & 0 & 0 \\ 0 & 0 & -1 & 0 \\ \sin(\theta_4) & \cos(\theta_4) & 0 & 0 \\ 0 & 0 & 0 & 1 \end{bmatrix} \tag{5} $$

$$ {}^{4}T_5 = \begin{bmatrix} \cos(\theta_5) & -\sin(\theta_5) & 0 & 0 \\ 0 & 0 & 1 & -l_2 \\ -\sin(\theta_5) & -\cos(\theta_5) & 0 & 0 \\ 0 & 0 & 0 & 1 \end{bmatrix} \tag{6} $$

$$ {}^{5}T_6 = \begin{bmatrix} \sin(\theta_6) & \cos(\theta_6) & 0 & 0 \\ 0 & 0 & -1 & 0 \\ -\cos(\theta_6) & \sin(\theta_6) & 0 & 0 \\ 0 & 0 & 0 & 1 \end{bmatrix} \tag{7} $$

$$ {}^{6}T_7 = \begin{bmatrix} \cos(\theta_7) & -\sin(\theta_7) & 0 & a_1 \\ 0 & 0 & 1 & 0 \\ -\sin(\theta_7) & -\cos(\theta_7) & 0 & 0 \\ 0 & 0 & 0 & 1 \end{bmatrix} \tag{8} $$

$$ {}^{0}_{7}T = \left[ {}^{0}_{1}T\; {}^{1}_{2}T\; {}^{2}_{3}T\; {}^{3}_{4}T\; {}^{4}_{5}T\; {}^{5}_{6}T\; {}^{6}_{7}T \right] \tag{9} $$

Lagrange's energy formulation was used to derive the equations of motion of the exoskeleton robot. The system potential energy is expressed as a function of the generalized coordinates, as shown in (10)-(11), whereas the kinetic energy is defined in terms of the generalized velocities in (12)-(13). The dynamic equations of motion were obtained by applying Lagrange's equations to the total system energy, as given in (14).

$$ u_i = -m_i .\ {}^{0}g^{T} .\ {}^{0}P_{ci} + u_{ref} \tag{10} $$

$$ u = \sum_{i=1}^{n} u_i \tag{11} $$

$$k_i = \left[\frac{1}{2} m_i . v_{ci}^T . v_{ci} + \frac{1}{2} \, {}^{i}\omega_{ci}^T . \, {}^{ci}I_i . \omega_{ci}\right] \tag{12}$$

In Equation (10), $m_i$ is the mass of the link, $g$ is the gravitational acceleration, and ${}^0P_{ci}$ is the location of the center of mass of the link with respect to the reference ground. In Equation (11), $n$ is the total number of links. In Equation (12), $m_i$ present the mass of the link, $v_{ci}$ and $\omega_{ci}$ express the linear and angular velocity of the link at the link's center of mass and $I_i$ is the link's moment of inertia at its center of mass.

$$k = \sum_{i=1}^{n} k_i \tag{13}$$

$$\tau_i = \left[\frac{d}{dt}\frac{\delta k}{\delta \dot{\theta}_i} - \frac{\delta k}{\delta \theta_i} + \frac{\delta u}{\delta \theta_i}\right] \tag{14}$$

The lower limb anthropometric parameters that were used for developing the robot dynamics equation of motion are given in Equation (15)-(35),

$$B_d = 0.6905 + 0.0297C \frac{lb}{ft^3}, where\ C = HW^{-\frac{1}{3}}, \tag{15}$$

In Equation (15), $B_d$ is the body density ($lbs/ft^3$), $H$ is the height of the subject ($inch$), $W$ is the body weight of the subject ($lbs$). Thigh, shank, and foot densities determined by Equation (16) to (18)

$$T_d = 1.035 + 0.814 * B_d \frac{lb}{ft^3} \tag{16}$$

$$S_d = 1.065 + B_d \frac{lb}{ft^3} \tag{17}$$

$$F_d = 1.071 + B_d \frac{lb}{ft^3} \tag{18}$$

In Equition (16)-(18), $T_d$ is the thigh density, $S_d$ is the shank density and $F_d$ is the foot density. The whole-body volume ($B_v$) was calculated using body weight and body density (Equation (19)),

$$B_v = \frac{W}{B_d} ft^3 \tag{19}$$

The volume of the thigh, shank, and foot were calculated as follows (Equation. (20) to Equation. (22))

$$T_v = 0.0922 * B_v \, ft^3 \tag{20}$$

$$S_V = 0.0464 * B_v \, ft^3 \tag{21}$$

$$F_v = 0.0124 * B_v \, ft^3 \tag{22}$$

The weights of the thigh, shank, and foot were calculated by the Equation (23)-(25).

$$T_m = T_v * T_d \, lbs \tag{23}$$

$$S_m = S_v * S_d \; lbs \tag{24}$$

$$F_m = F_v * F_d \; lbs \tag{25}$$

The length of the thigh ($T_l$), shank ($S_l$), foot ($F_l$) and ankle to the lower face of the foot ($A_g$) were calculated by Equation (26)-(29),

$$T_l = 0.245 * H \; inch, \tag{26}$$

$$S_l = 0.285 * H \; inch, \tag{27}$$

$$F_l = 0.152 * H \; inch, \tag{28}$$

$$A_g = 0.043 * H \; inch \tag{29}$$

The locations of the center of the mass from the proximal joint (for thigh ($T_{cm}$), shank ($S_{cm}$), foot ($F_{cm}$)) are given by Equation (30)-(32).

$$T_{cm} = 0.41 * T_l \; inch, \tag{30}$$

$$S_{cm} = 0.393 * S_l \; inch, \tag{31}$$

$$F_{cm} = 0.445 * F_l \; inch \tag{32}$$

The empirical equations for the inertial properties of the thigh $T_i$, shank $S_i$, and foot $F_i$ are given in Equation (33)-(35),

$$T_i = \begin{bmatrix} T_m\,(0.124 * T_l)^2 & 0 & 0 \\ 0 & T_m(0.267 * T_l)^2 & 0 \\ 0 & 0 & T_m(0.267 * T_l)^2 \end{bmatrix} \tag{33}$$

$$S_i = \begin{bmatrix} S_m(0.281 * S_l)^2 & 0 & 0 \\ 0 & S_m(0.114 * S_l)^2 & 0 \\ 0 & 0 & S_m(0.275 * S_l)^2 \end{bmatrix} \tag{34}$$

$$F_i = \begin{bmatrix} F_m(0.124 * F_l)^2 & 0 & 0 \\ 0 & F_m(0.245 * F_l)^2 & 0 \\ 0 & 0 & F_m(0.257 * F_l)^2 \end{bmatrix} \tag{35}$$

Equation (15)-(35) were incorporated into the dynamic equation of motion of the robot.

The robot's dynamic equation of motion ((14) can be rewritten as Equation (36),

$$\tau_{Joint} = \left[M(\theta)\ddot{\theta} + V\left(\theta, \dot{\theta}\right) + G(\theta)\right] \tag{36}$$

A 7 DOF human lower extremity dynamic model was developed. In Equation (36), $M(\theta)$ is the mass matrix which is a $(7x7)$ symmetric positive definite matrix , $V(\theta, \dot{\theta})$, $(7x1)$ is the Coriolis and the centripetal term, and the gravitational term is represented by $G(\theta)$, $(7x1)$matrix. $\tau_{Joint}$, $(7x1)$ presents the joints torque requirements.

The robot dynamic equation of motion Eqn. (36) can be written as,

$$\ddot{\theta} = M(\theta)^{-1}[\tau_{Joint} - V(\theta, \dot{\theta}) - G(\theta)] \quad (37)$$

Equation 37 pictorially presented by Figure 2 which represents the architecture of an ideal robot dynamics (without considering joint friction).

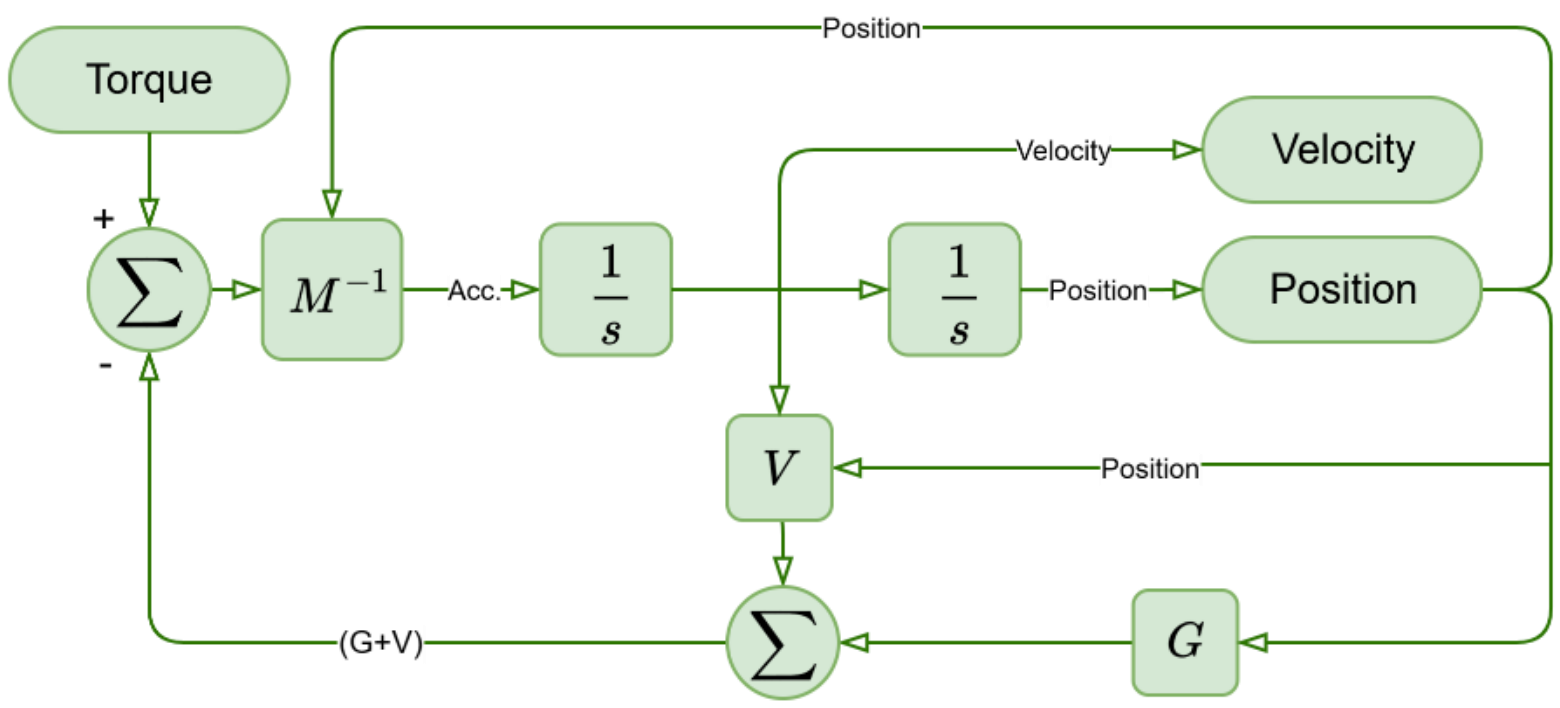

Figure 2 Internal architecture of the robot model

Naturally, friction is induced between two mechanical mating parts with relative motion. The robot dynamics grow more complex by incorporating joint friction torques.

$$\tau_{Joint} = [M(\theta)\ddot{\theta} + V(\theta, \dot{\theta}) + G(\theta) + \tau_{friction}] \quad (38)$$

Where,

$$\tau_{friction} = [f(\dot{\theta})] \quad (39)$$

Eqn. (37) can be rewritten as Eqn. (40)

$$\ddot{\theta} = M(\theta)^{-1}[\tau_{Joint} - V(\theta, \dot{\theta}) - G(\theta) - f(\dot{\theta})] \quad (40)$$

The robot dynamics with frictional disturbances are schematically presented in Figure 3

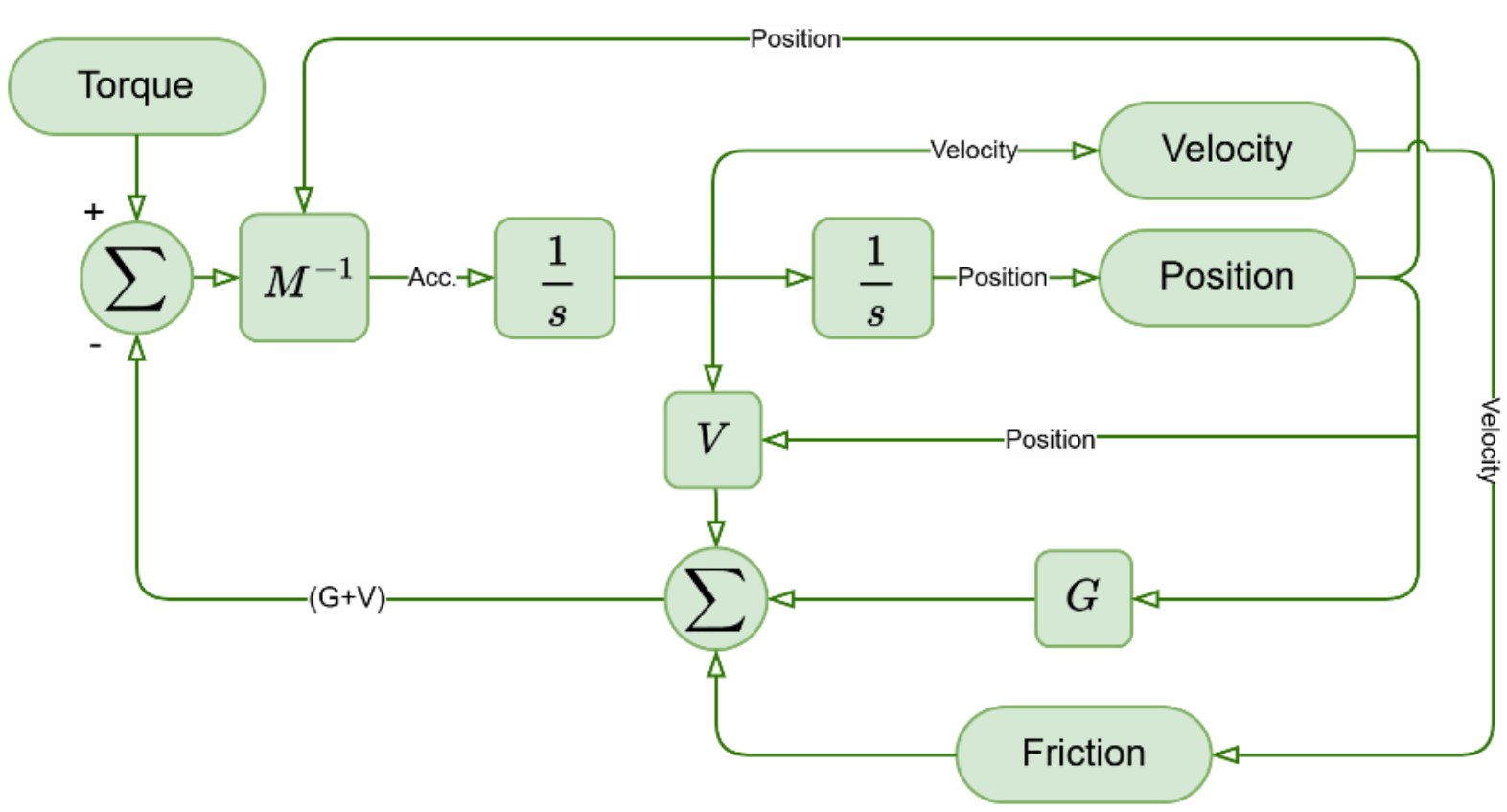

Figure 3 Internal architecture of the physical Robot model

The used friction model combined the Coulomb friction, the Viscous friction, and the Stribeck effect [38].

The Equation. (41) to Equation. (43) approximate the joint friction torques:

$$T = \sqrt{(2e)}(T_{brk} - T_C).\exp\left(-\left(\frac{\omega}{\omega_{St}}\right)^2\right).\frac{\omega}{\omega_{St}} + T_C.\tanh\left(\frac{\omega}{\omega_{Coul}}\right) + f\omega \tag{41}$$

$$\omega_{St} = \omega_{brk}\sqrt{2} \tag{42}$$

$$\omega_{Coul} = \frac{\omega_{brk}}{10} \tag{43}$$

Where,

$\boldsymbol{T}$ represents the total friction

$\boldsymbol{T_C}$ represents the Coulomb friction

$\boldsymbol{T_{brk}}$ represents the breakaway friction torque: The Breakaway friction is defined as the sum of the Coulomb and Stribeck frictions in the vicinity of zero velocity.

$\boldsymbol{\omega_{brk}}$ represents the breakaway friction velocity: The velocity at which the Stribeck friction is at its peak. At this point, the sum of the Stribeck and Coulomb friction is the Breakaway friction force.

$\boldsymbol{\omega_{St}}$ represents the Stribeck velocity threshold

$\boldsymbol{\omega_{Coul}}$ represents the Coulomb velocity threshold

$\boldsymbol{\omega}$ represents the input angular velocity

$\boldsymbol{f}$ is the coefficient of viscous friction: The coefficient of proportionality between the friction torque and the angular velocity. The parameter must have a positive value.

The simulation of the developed friction model is shown in Figure 4.

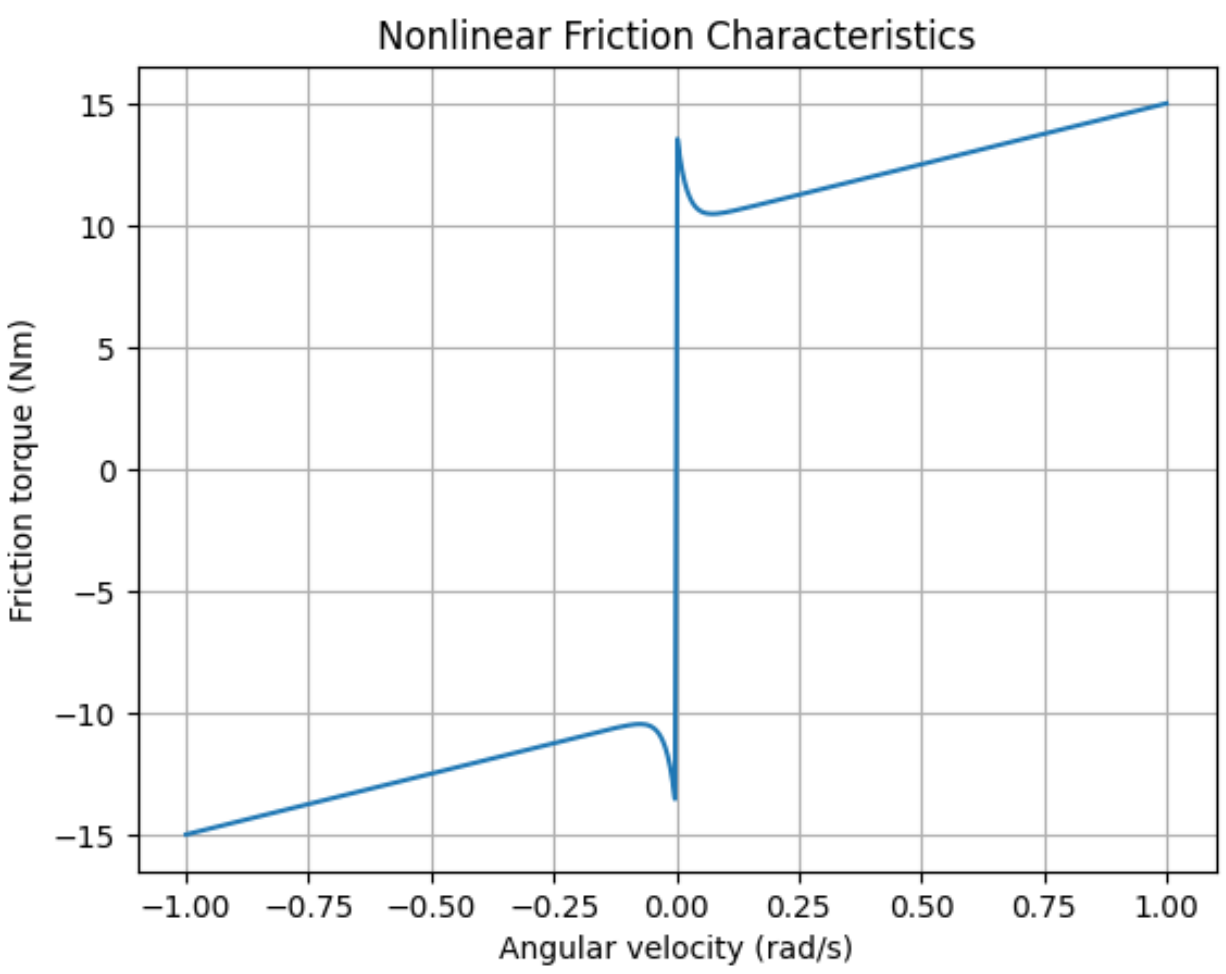

Figure 4 Friction model simulation

The parameters used for friction simulation (Figure 4) are given below,

$T_{Peak} = 100\ Nm, \omega = 100\ rad/sec, f = 5\ Nm/(rad/sec), T_{Columb} = 0.1 * T_{Peak}\ Nm,$

$\omega_{brk} = 0.01\ rad/sec, T_{brk} = 0.15 * T_{Peak}\ Nm$

The following section will discuss the Computed Torque control scheme.

## 1. Computed Torque control

The computed torque control scheme calculates robot joints' trajectory tracking torque requirements by using robot inverse dynamics. The computed torque control system accomplishes two goals: linearize the nonlinear robot dynamics by providing the required torque to mitigate the effects of gravity and Coriolis-centrifugal force, remove the effects of position-dependant mass or inertia (Figure 5, linearization loop) and to deliver the controlled torque to track the robot input trajectories (Figure 5, control loop).

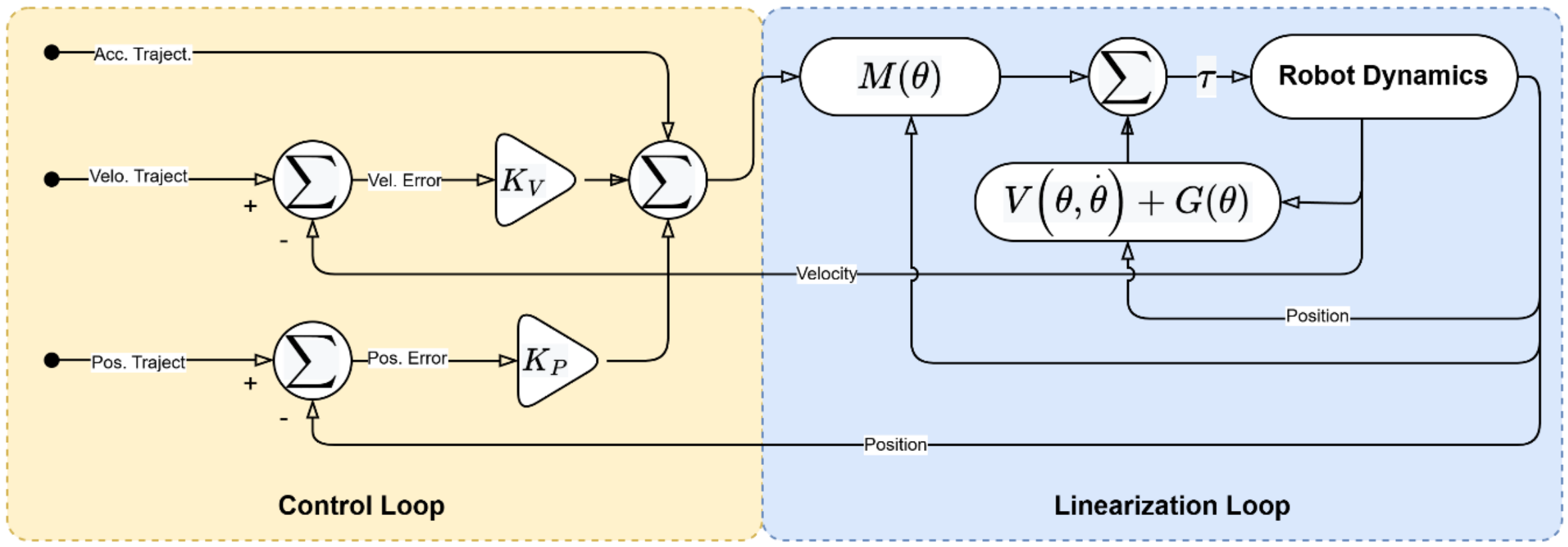

Figure 5 Computed torque control scheme

The main drawbacks of a computed torque controller are: an accurate dynamic model is impractical and the control algorithm involves execution of the whole robot dynamics on every sample time. It's a torque control scheme that requires a higher sampling rate compared to the position or velocity control scheme. According to the robot computed torque control scheme, the input torque to the robot is given by Equation (44),

$$\tau = M(\theta)\left[\ddot{\theta}_d + K_v\left(\dot{\theta}_d - \dot{\theta}\right) + K_p(\theta_d - \theta)\right] + V\left(\theta, \dot{\theta}\right) + G(\theta) \qquad (44)$$

In Equation (44), $\theta_d$ , $\dot{\theta}_d$ and $\ddot{\theta}_d$ presents the robot's desired trajectories. The actual position, velocity, and acceleration of the robot are represented by $\theta, \dot{\theta}$ and $\ddot{\theta}$. $K_p, K_v$ are the positive definite gain matrices that act as the PD controller gain. $M(\theta)$ presents the robot mass matrix, $V(\theta, \dot{\theta})$ presents the Coriolis and centrifugal force and $G(\theta)$ is the gravitational term of the robot dynamics.

## 2. Dynamic modeling using Deep Neural Network

A neural network consists of a set of interconnected artificial neurons organized in layers. Neural networks containing more than three layers are commonly referred to as deep neural networks (DNNs) [39]. DNNs have gained significant attention as powerful modeling frameworks, including applications that emulate neural computations observed in the primate brain [40]. A key advantage of DNNs over shallow architectures is their ability to learn hierarchical and abstract feature representations from data [41]. This layered structure enables DNNs to efficiently capture complex nonlinear relationships that are difficult to model using conventional machine learning techniques [42].

The increased depth of DNNs allows for the extraction of intricate and high-level representations of input data, making them well suited for solving problems that are challenging or intractable for traditional learning approaches [43]. While the presence of multiple layers increases computational complexity, the performance benefits of deep architectures typically outweigh this added burden [39]. Moreover, DNNs inherently support parallel computation, allowing efficient implementation on multicore CPUs, GPUs, or digital signal processors (DSPs). As a result, DNN-based control algorithms can often be executed more efficiently than conventional model-based control methods [44].

In supervised learning, labeled training data are used to train a neural network to predict outputs for previously unseen inputs. For multilayer neural networks, the supervised learning process is typically carried out through forward propagation to compute network outputs and backward propagation to update network parameters by minimizing a defined loss function.

In the forward propagation process, the input layer of the deep neural network receives a feature vector consisting of robot input trajectories along with user-specific parameters such as weight and height, resulting in a total of 23 input variables. The network architecture includes three fully connected hidden layers with 21, 14, and 7 neurons, respectively. At each neuron, inputs from the previous layer are multiplied by their associated weights, summed with a bias term, and passed through a nonlinear activation function to generate the neuron output. Hyperbolic tangent activation functions are employed in the hidden layers to introduce nonlinearity, while a pure linear activation function is used at the output layer. The output layer comprises seven neurons, each producing a continuous-valued prediction corresponding to joint torque outputs. This configuration enables the network to model complex nonlinear mappings between the input features and the desired torque outputs. Figure 7 illustrates the hyperbolic tangent and pure linear activation functions.

Following forward propagation, supervised learning proceeds through backpropagation to update the network parameters. The training process begins by computing the loss between the predicted outputs and the target values using the mean squared error (MSE) function, which is appropriate for the regression problem considered in this study. The gradient of the loss function with respect to the output layer activations is first evaluated and then propagated backward through the network layers. Using the chain rule of calculus, gradients with respect to the weights and biases in each layer are computed recursively. These gradients are subsequently used to update the network parameters in order to minimize the loss function. In the developed deep neural network, the Levenberg–Marquardt optimization algorithm is employed to update the weights and biases efficiently. This iterative process of forward propagation, loss evaluation, and backward propagation is repeated across the training dataset, enabling the network to progressively improve its prediction accuracy.

The primary objectives in designing the deep neural network architecture were to maximize prediction accuracy, minimize overfitting, and reduce computational complexity to ensure real-time feasibility on the physical controller. The developed multilayer neural network architecture [45] is illustrated in Figure 6 Multiple network configurations were systematically evaluated, and the architecture shown in Figure 7 yielded the optimal balance between prediction performance and generalization capability.

The finalized deep neural network consists of a total of 49 neurons distributed across three hidden layers and one output layer, arranged as $L_1 \times 21$, $L_2 \times 14$, $L_3 \times 7$, and $L_4 \times 7$. The use of multiple hidden layers, rather than a single hidden layer, enables the network to learn higher-level abstract features while improving generalization and reducing susceptibility to overfitting. The complete network contains 224 trainable parameters, including both weights and biases. Hyperbolic tangent (tan-sigmoid) activation functions were employed in the hidden layers, whereas a pure linear activation function was applied at the output layer to support continuous torque prediction. Several activation functions were evaluated during network development, and the tan-sigmoid function demonstrated superior performance in terms of prediction accuracy and overfitting mitigation. It was also observed that increasing the number of neurons in any layer or introducing additional layers led to overfitting without improving predictive performance.

The neural network was implemented using the MATLAB feedforwardnet(hiddenSizes, trainFcn) function [46]. During supervised training, the Levenberg-Marquardt optimization algorithm was used to update the network weights and biases during backpropagation, owing to its fast convergence and robustness for nonlinear regression problems [47, 48].

For training the developed neural network, two distinct datasets were generated. The first dataset corresponds to sequential joint movements, in which the seven joints are actuated successively, while the second dataset represents simultaneous joint movements, where all joints initiate motion concurrently. Training data were generated through hierarchical combinations of robot joint positions, velocities, and accelerations, along with subject-specific parameters including height and weight. The hierarchical structure of the training data generation process is illustrated in Figure 8.

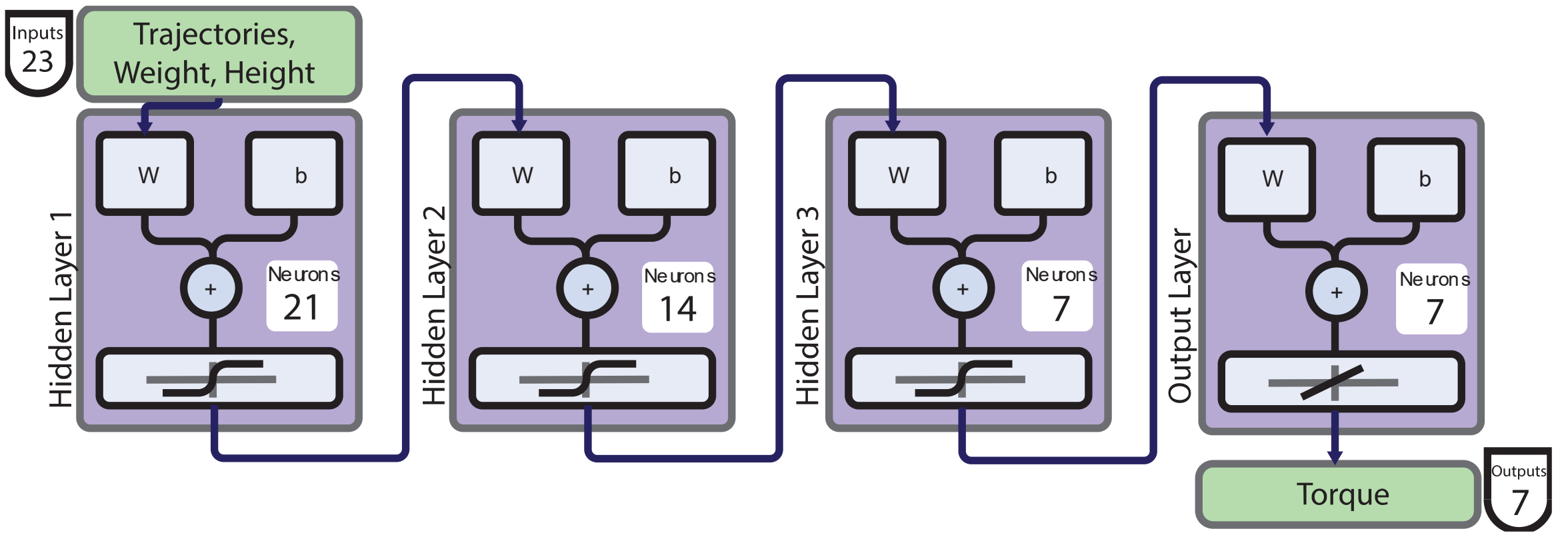

Figure 6 Architecture of the developed deep neural network

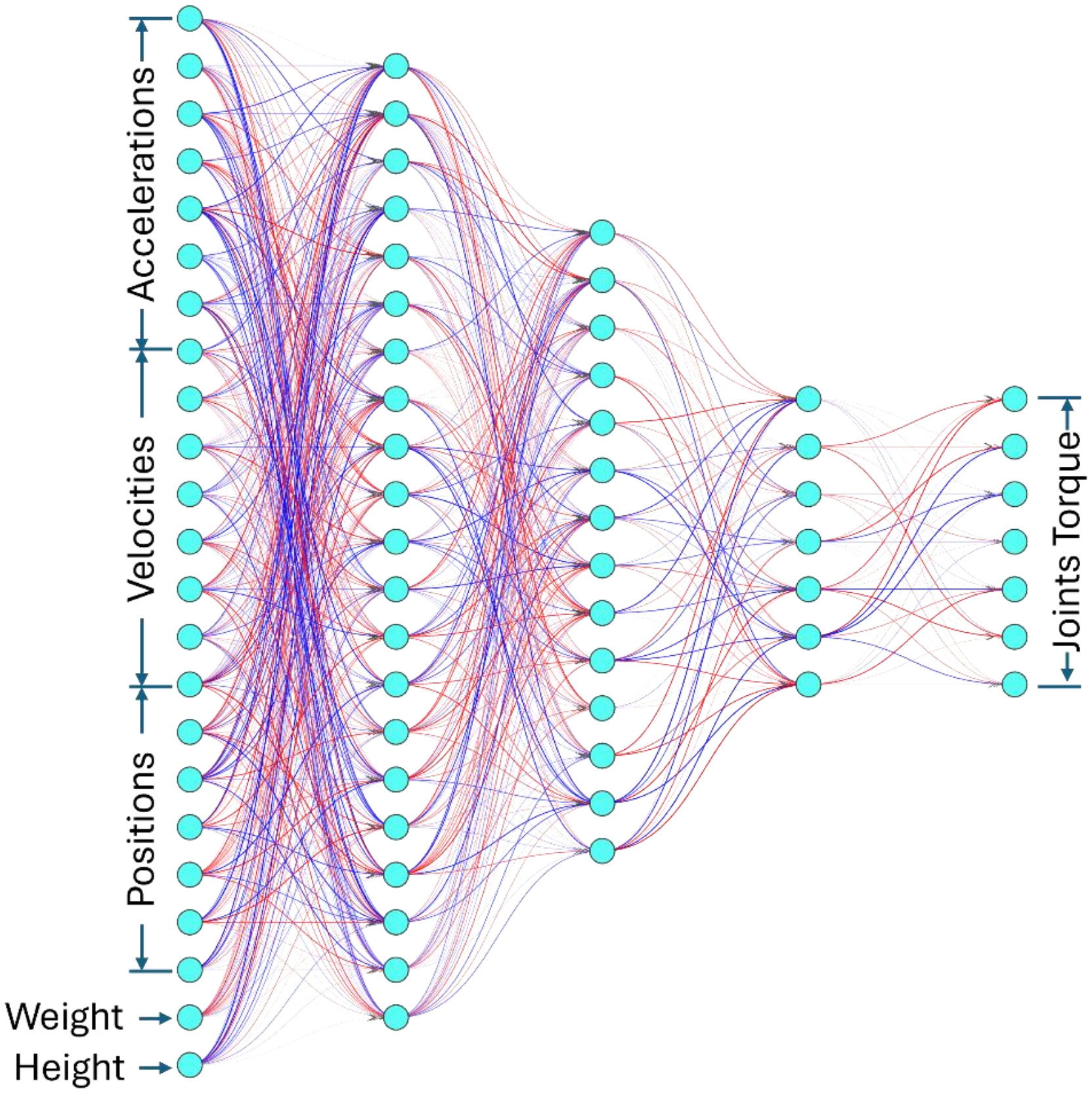

Figure 7 Developed deep neural network for joint torque estimation

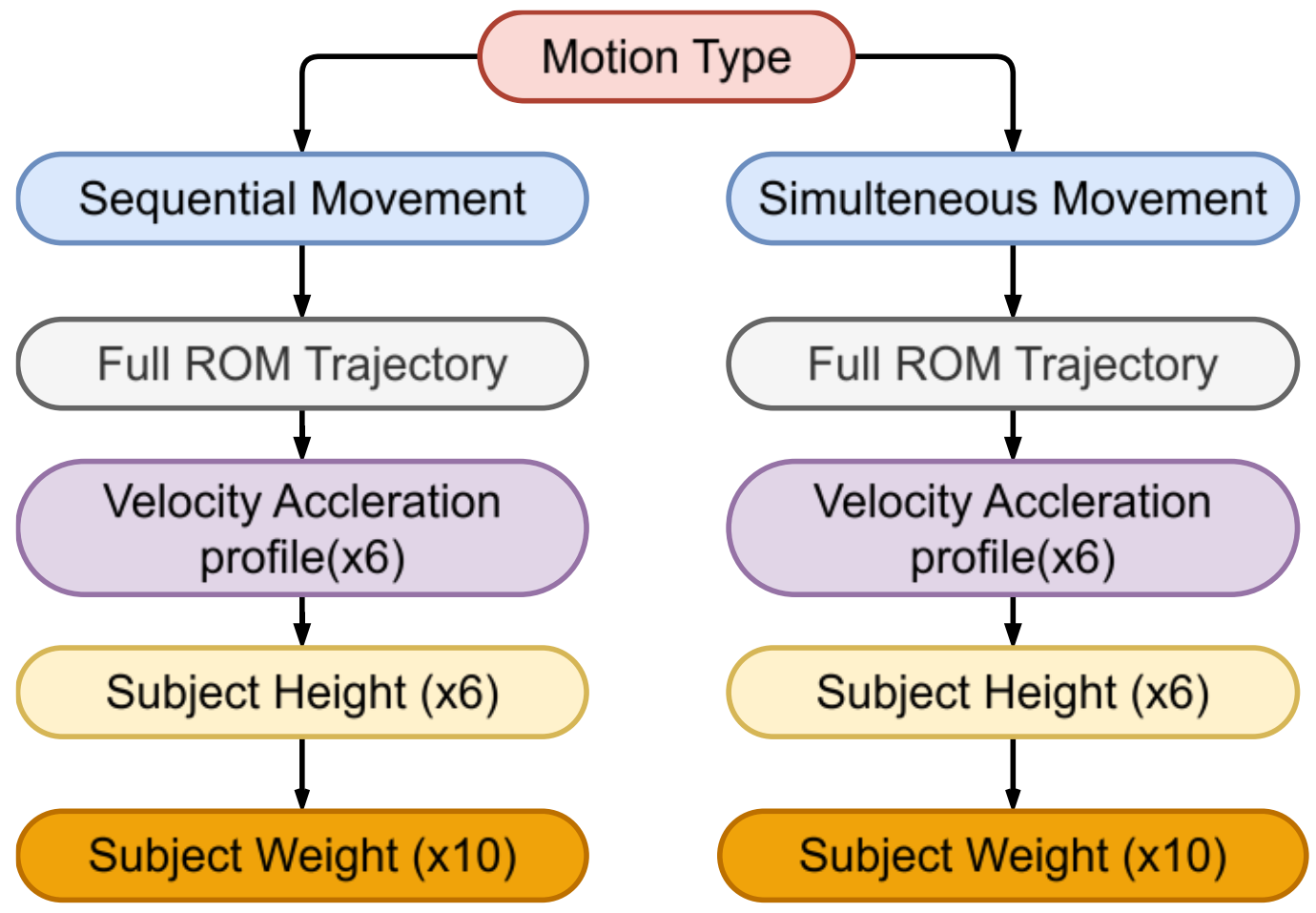

Figure 8 Hierarchical data generation schematic diagram

Seven joint positions, velocities, and accelerations, together with subject-specific parameters including weight and height obtained from various combinations, were applied as inputs to the computed-torque–controller-based dynamic simulation. The corresponding outputs, consisting of the seven joint torque signals generated by the computed torque controller, were recorded and used as training data for the neural network. Figure 9 illustrates the schematic diagram of the computed-torque controller-based dynamic simulation, where the input signals are highlighted in purple and the output torque signals are shown in green.

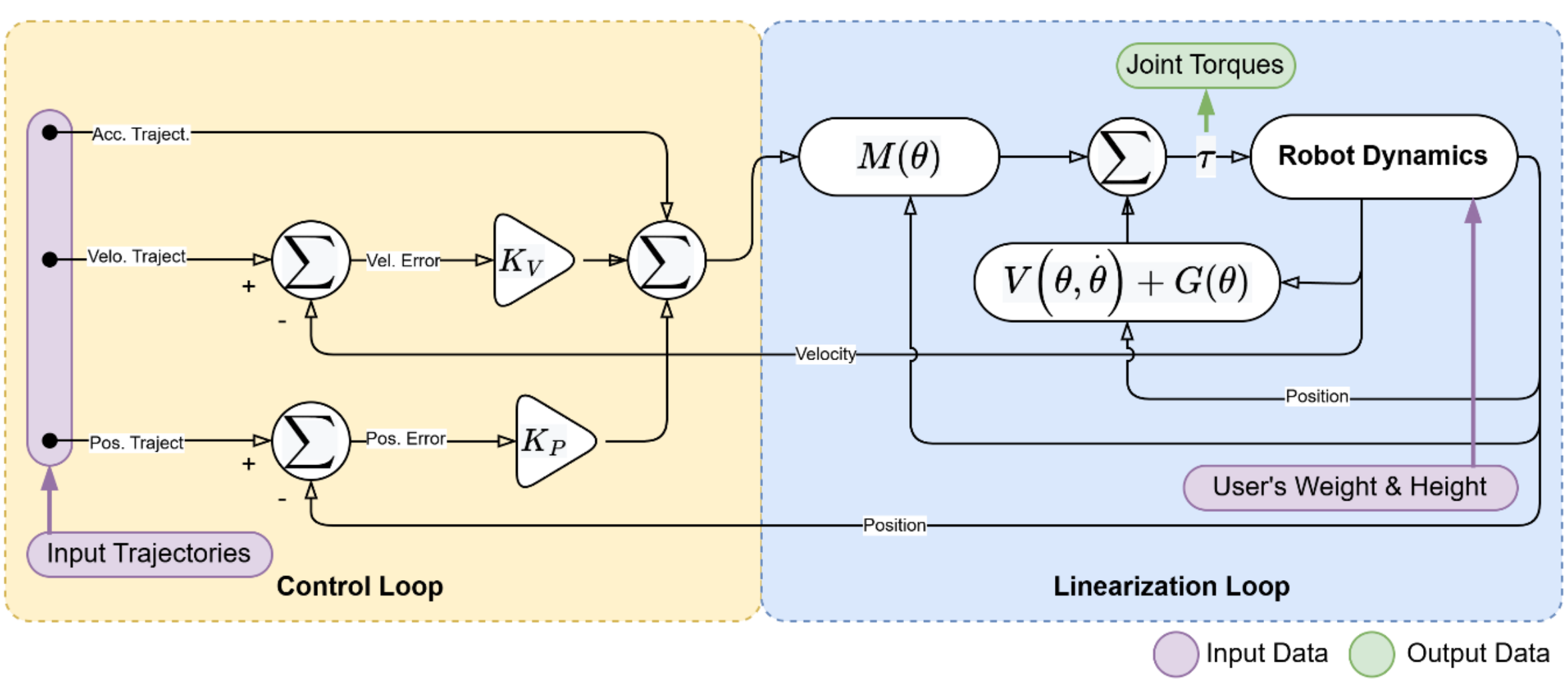

Figure 9 Input and Output data tracing locations in the simulation

Simulations were conducted for both sequential and simultaneous joint movements, yielding a total of 43,560,880 data samples. Due to the large data volume, the available server workstation equipped with a 32-core Intel Xeon processor and 48 GB of memory was unable to process the complete dataset efficiently.

Consequently, the collected data were downsampled by a factor of ten to reduce the computational burden and storage requirements. Correlation analysis was subsequently performed to identify the most influential input features and potentially reduce the network input dimensionality. The analysis indicated that the input trajectories, subject weight, and height exhibited strong correlation with the resulting joint torques; therefore, all input features were retained.

Following downsampling, a total of 4,356,088 data samples were used for neural network training. Each training sample comprised 23 input variables, including the desired positions, velocities, and accelerations of the seven joints, along with the subject's height and weight, and seven output variables corresponding to the predicted joint torques.

For the generation of the input dataset, full ranges of motion of the human lower extremity joints were considered. Figure 10 illustrates the corresponding range-of-motion trajectories, where the left subfigure depicts simultaneous joint movements and the right subfigure shows sequential joint movements.

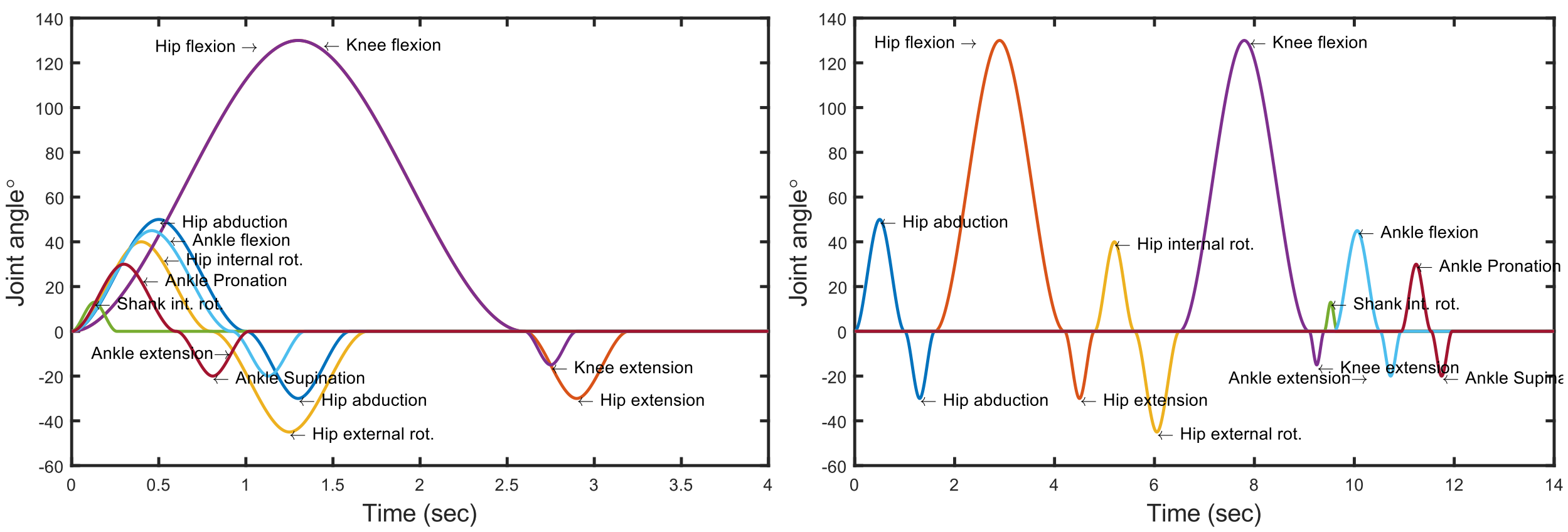

Figure 10 Simultaneous and sequential joint movements (constructed based on the human lower extremity's full ranges of motion [14])

To generate the training data for the deep neural network, full range-of-motion position trajectories were executed at multiple velocity profiles while considering subjects with different heights and body masses. The various configurations employed for training data generation are summarized in Table 2.

Table 2 Different values of joints velocities, subject's weight and height used for training data generation

| | |
|---|---|
| Joints velocity | 10, 20, 30, 40, 50, 60, 70, 80, 90, 100 [deg/sec] |
| Subject's height | 50, 55, 60, 70, 75 [inch] |
| Subject's weight | 150, 160, 170, 180, 190, 200, 210, 220, 230, 240, 250 [lbs.] |

The primary function of the developed deep neural network is to estimate joint torque requirements based on the input motion trajectories and subject-specific parameters, namely height and weight. During training, 70% of the available dataset was allocated for network training, 15% for validation, and the remaining 15% for testing. Training performance was evaluated using the mean squared error (MSE) metric. Figure 11 illustrates the training, validation, and testing performance of the developed neural network.

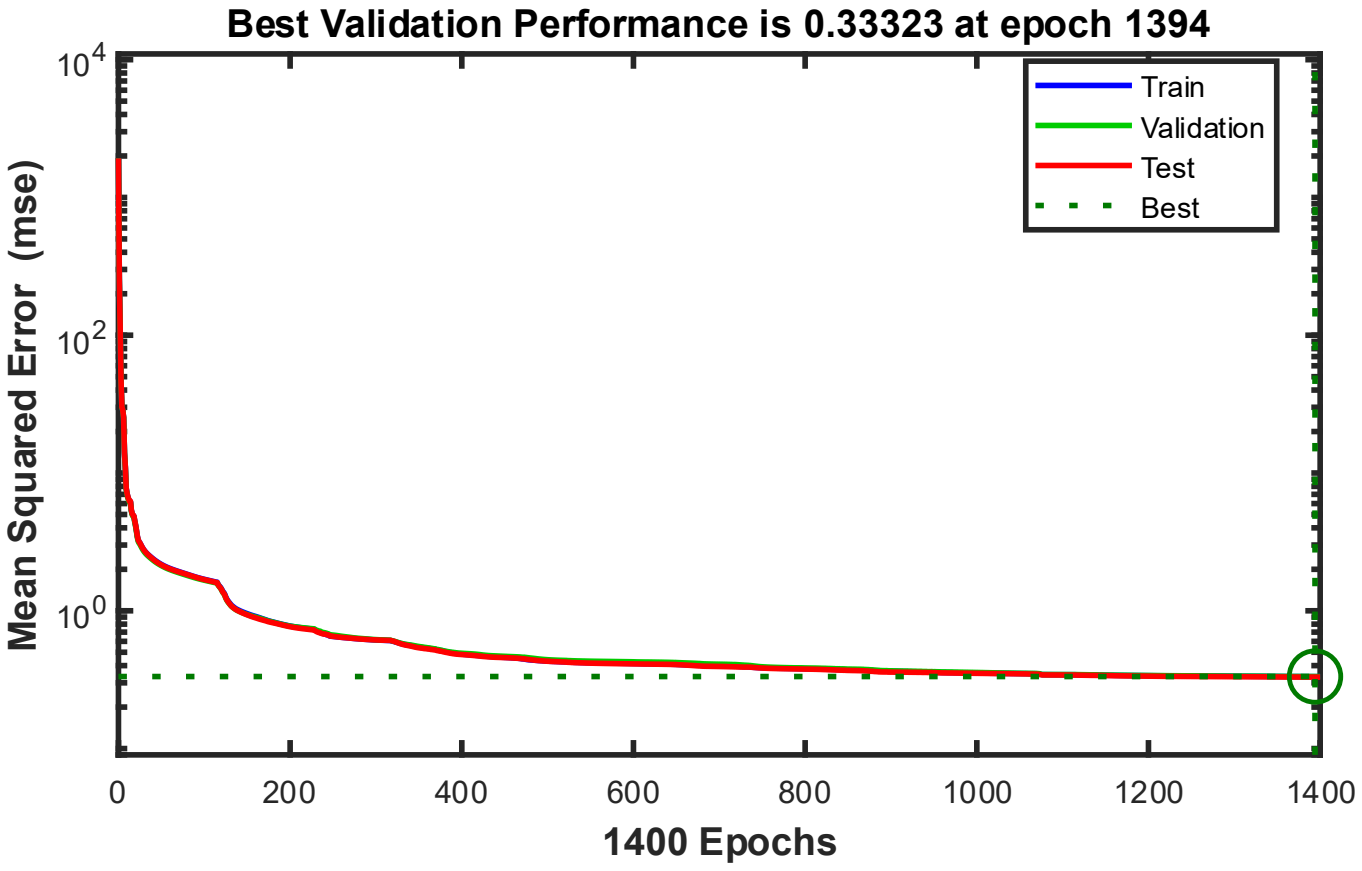

Figure 11 Training Neural network

## 3. Control algorithm based on the developed deep neural network

Figure 12 illustrates the architecture of the proposed hybrid controller developed to drive the robot along the specified reference trajectories. The reference motion trajectories, together with the subject's height and weight, are provided as inputs to the trained deep neural network, which predicts the required torques for the seven joints. The deep neural network output is utilized as a feedforward component within the control architecture. To compensate for prediction errors and modeling uncertainties, a proportional-derivative (PD) controller is incorporated in the feedback loop. Uniform proportional and derivative gains of $K_P = 3000$ and $K_V = 250$, respectively, were applied to all joints of the PD controller.

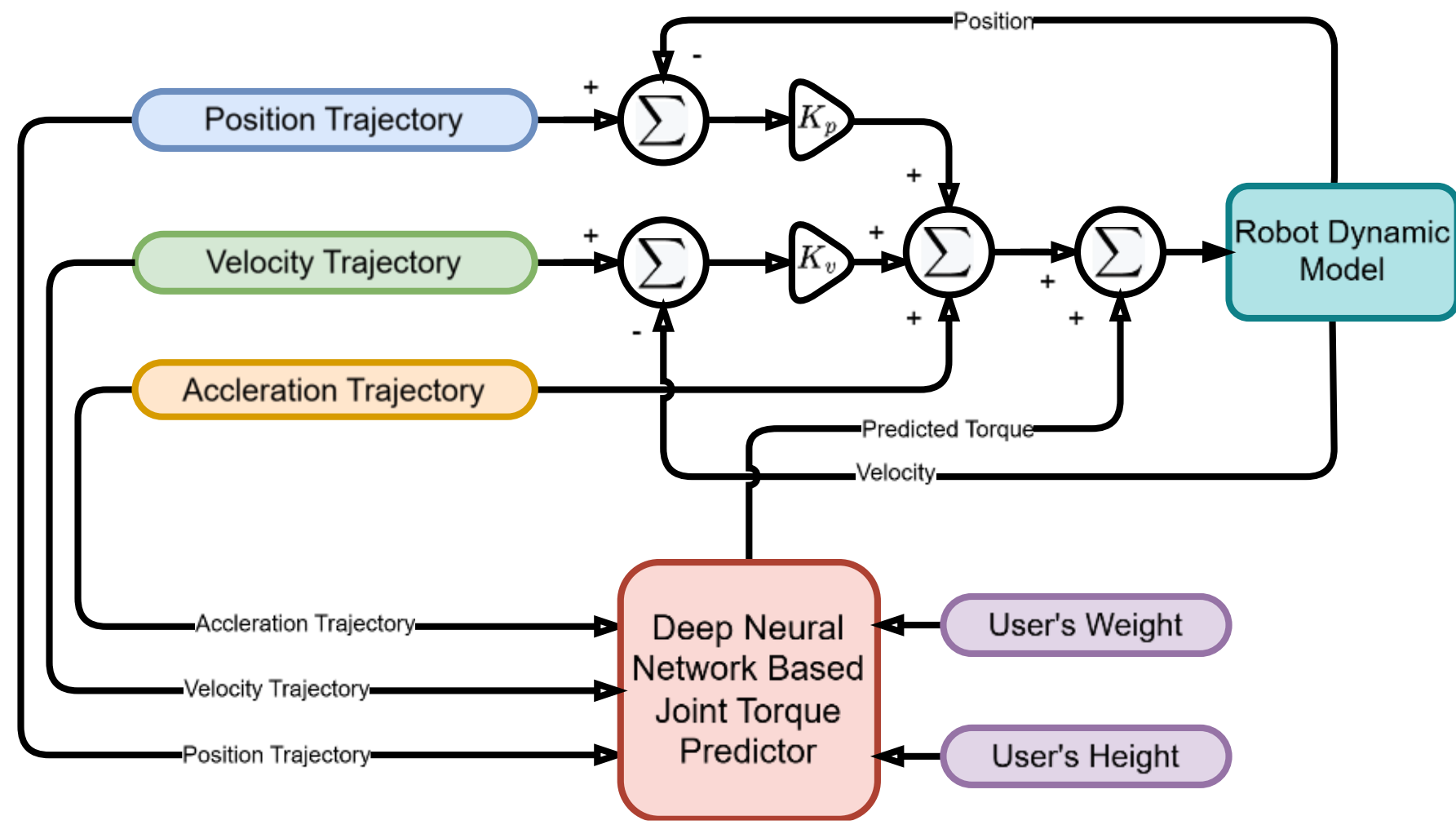

Figure 12 Control architecture of the Deep neural network-based robot controller

The prediction accuracy of the developed deep neural network depends strongly on the representativeness and diversity of the training data, while overall performance is assessed through validation and testing results. The training dataset was generated using the analytical dynamic model of the exoskeleton robot in combination with reference input trajectories and subject-specific parameters, namely height and weight. To ensure high predictive accuracy and robust generalization, a sufficiently large dataset was created by systematically varying these parameters across multiple combinations.

High prediction accuracy of the deep neural network enables effective feedforward compensation by accurately approximating the nonlinear system dynamics, including gravitational, Coriolis, and centrifugal effects, as well as the position-dependent inertia matrix. Consequently, the trained deep neural network serves as a feedforward controller within the hybrid control framework. A proportional-derivative (PD) controller is incorporated in the feedback loop to compensate for residual prediction errors and unmodeled dynamics. Figure 12 illustrates the overall control architecture of the proposed controller.

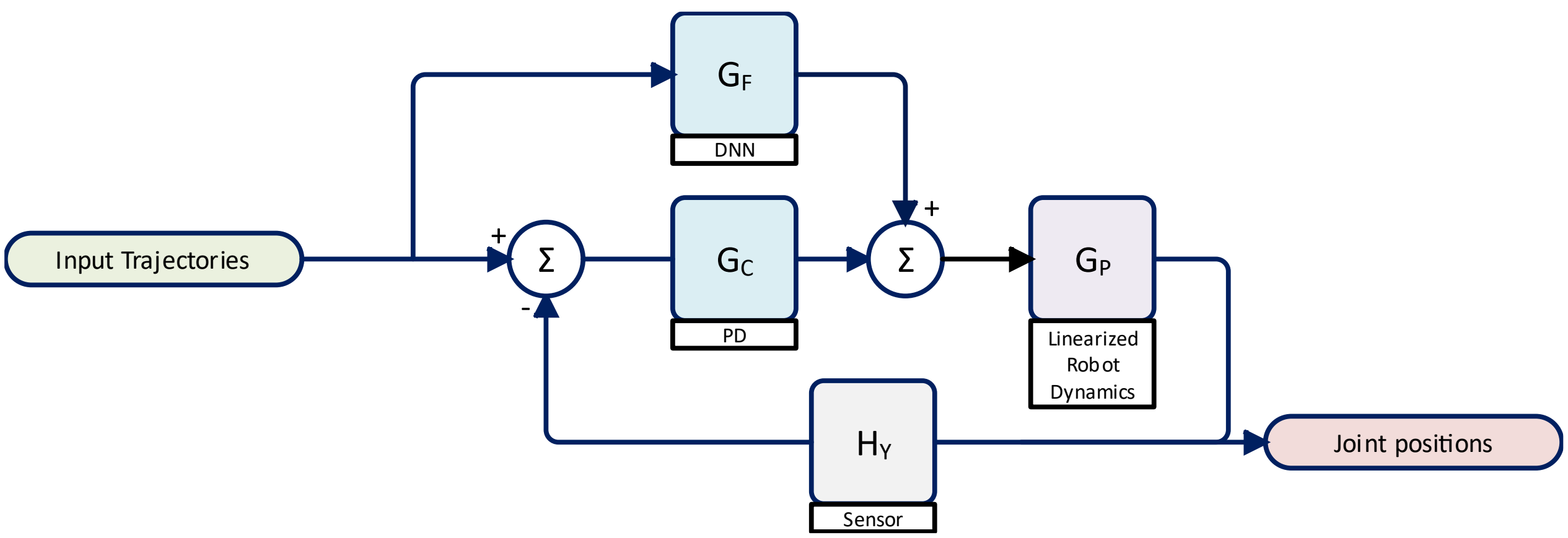

Figure 13 Control architecture of the hybrid controller

Figure 13 illustrates the control architecture of the developed hybrid controller. The closed-loop transfer function relating the input reference trajectories to the joint positions can be expressed as,

$$\frac{\text{Joint position}}{\text{Input trajectories}} = \frac{G_P \mathbf{G_F}(1 + G_C G_P H_Y) + G_C G_P}{1 + G_C G_P H_Y} \quad (45)$$

where $G_F$represents the feedforward gain, $G_C$ denotes the feedback controller, $G_P$is the plant dynamics, and $H_Y$is the feedback path transfer function. The corresponding characteristic equation of the closed-loop system is given by,

$$1 + G_C G_P H_Y = 0. \quad (46)$$

System stability is governed by the roots of the characteristic equation and is therefore independent of the feedforward gain $G_F$. The developed deep neural network provides effective feedforward compensation by eliminating the nonlinear dynamic components arising from gravitational, Coriolis, and centrifugal forces, as well as canceling the effects of the position-dependent mass matrix. Once these nonlinearities are compensated, the system dynamics are effectively linearized and decoupled, such that the joint angular accelerations become directly proportional to the applied torques, and the mass matrix behaves as a unity-gain proportional constant. Under these conditions, the resulting linearized and decoupled system dynamics are governed by the characteristic equation in (47).

$$\tau = \ddot{\theta} \quad (47)$$

By applying the Laplace transform

$$\tau(s) = s^2 \theta(s)$$

$$\frac{\theta(s)}{\tau(s)} = \frac{1}{s^2}$$

So, $G_p(s) = \frac{\theta(s)}{\tau(s)} = \frac{1}{s^2}$

PD controller can be written as, $G_c(s) = K_p + K_d s$

Due to higher bandwidth and unity gain of the sensor, transfer function can be considered as 1

$$H_y = 1$$

By substituting the values of $G_c, G_p, H_y$ into Eqn. (46). The characteristics equation become (48)

$$s^2 + K_d s + K_p = 0 \quad (48)$$

The constructed Routh Hurtz table based on the characteristic's equation (48) is given in Table 3

Table 3: Routh Hurtz table

| | | |
|---|---|---|
| $s^2$ | 1 | $K_p$ |
| $s^1$ | $K_d$ | |
| $s^0$ | $K_p$ | |

According to the Routh-Hurwitz stability criterion, the absence of sign changes in the first column of the Routh array guarantees system stability. As shown in Table 3, the proposed control system remains stable for all positive values of the proportional and derivative gains, $K_p$and $K_d$.

The stability of the developed controller does not depend on the feed-forward gain $G_F$, which is provided by the deep neural network. The feed-forward gain either amplifies or attenuates the output signal, it would not be able to propagate the trajectory tracking errors. A well trained deep neural network along with a PD controller with a positive gain ensures the stability of the developed hybrid controller. The performance of the deep neural network can be ensured during the training and validation phase. The smaller value of the mean square error during network training and validation indicates better prediction performance.

## 4. Simulation results analysis

In this section, the dynamic behavior of the human lower extremity is simulated using the developed deep neural network-based hybrid controller, and its trajectory tracking performance is evaluated. Figure 14 to Figure 19 present the tracking results for simultaneous joint movements. Figure 15 illustrates the trajectory tracking performance of all seven joints under simultaneous motion, where the reference trajectories and the corresponding output trajectories are plotted together. Owing to the very small magnitude of the tracking errors, the reference and actual trajectories are nearly indistinguishable in the figure.

The corresponding trajectory tracking errors are shown in Figure 15. The maximum observed tracking errors for joints 1 through 7 are $[0.26°, 0.20°, 0.20°, 0.26°, -0.14°, -0.21°, 0.19°]$, respectively. The simulation results indicate that joint 4 exhibits the highest tracking error of $0.26°$, followed by joints 1 and 6, while all joints maintain errors within a small and acceptable range, demonstrating the effectiveness of the proposed hybrid control strategy.

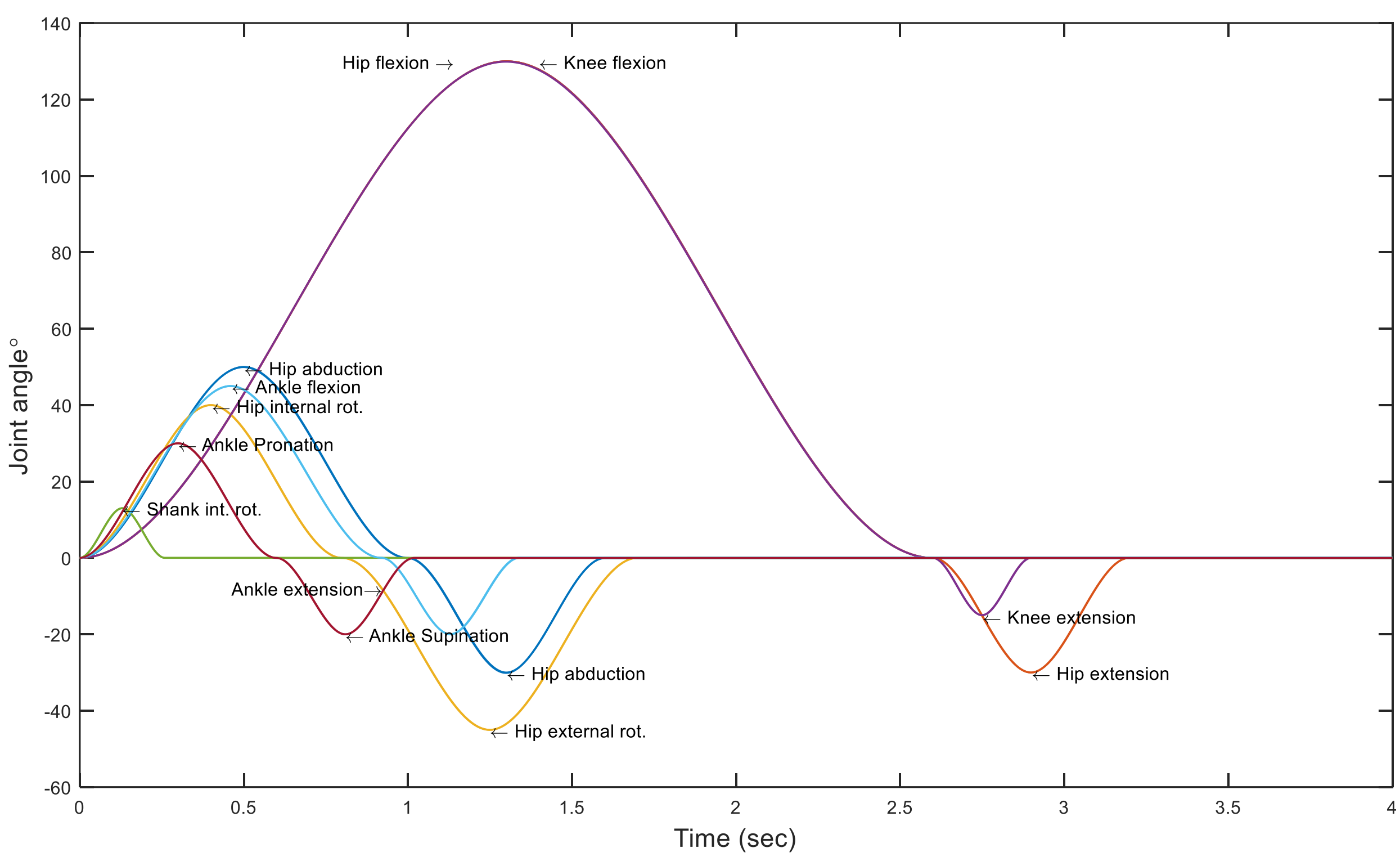

Figure 14 Trajectory tracking performance (Simultaneous joint movement)

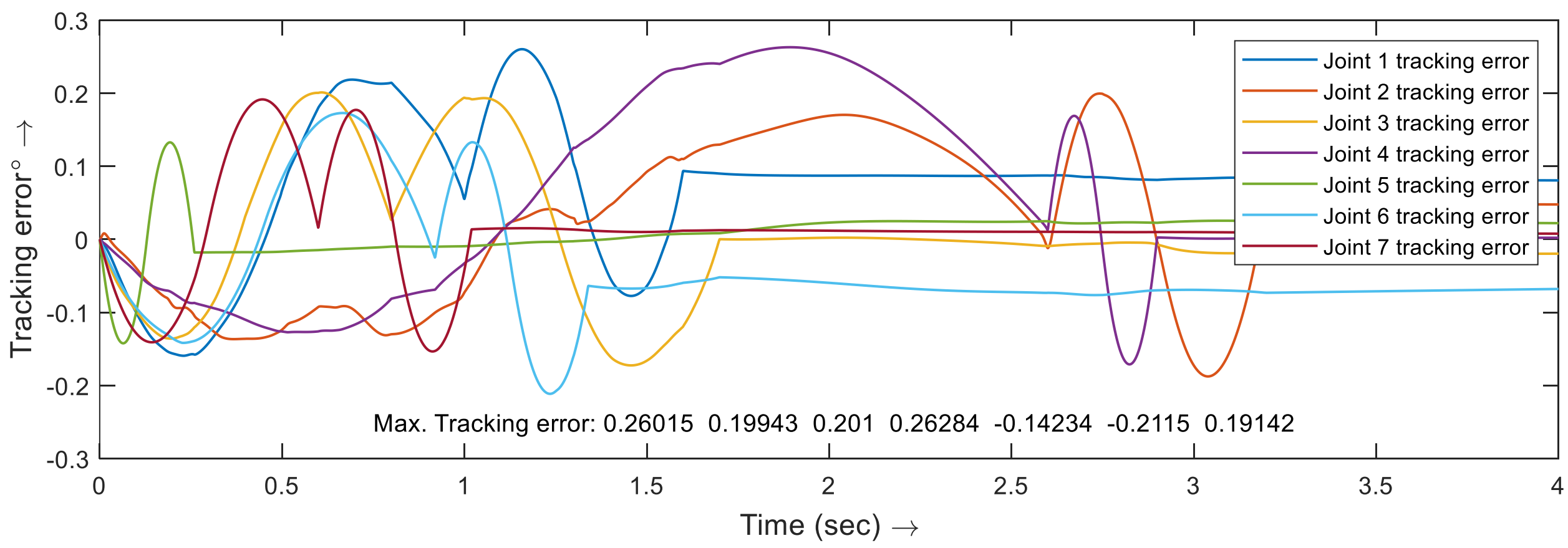

Figure 15 Trajectory tracking Error (Simultaneous joint movement)

Figure 16 shows the joint torque requirement for tracking the reference trajectories depicted in Figure 14. The left column of Figure 16 shows the total torque required for simultaneous trajectory tracking. The middle column of Figure 16 shows the predicted joints torques by the deep neural network and the right shows the contribution of the PD controller.

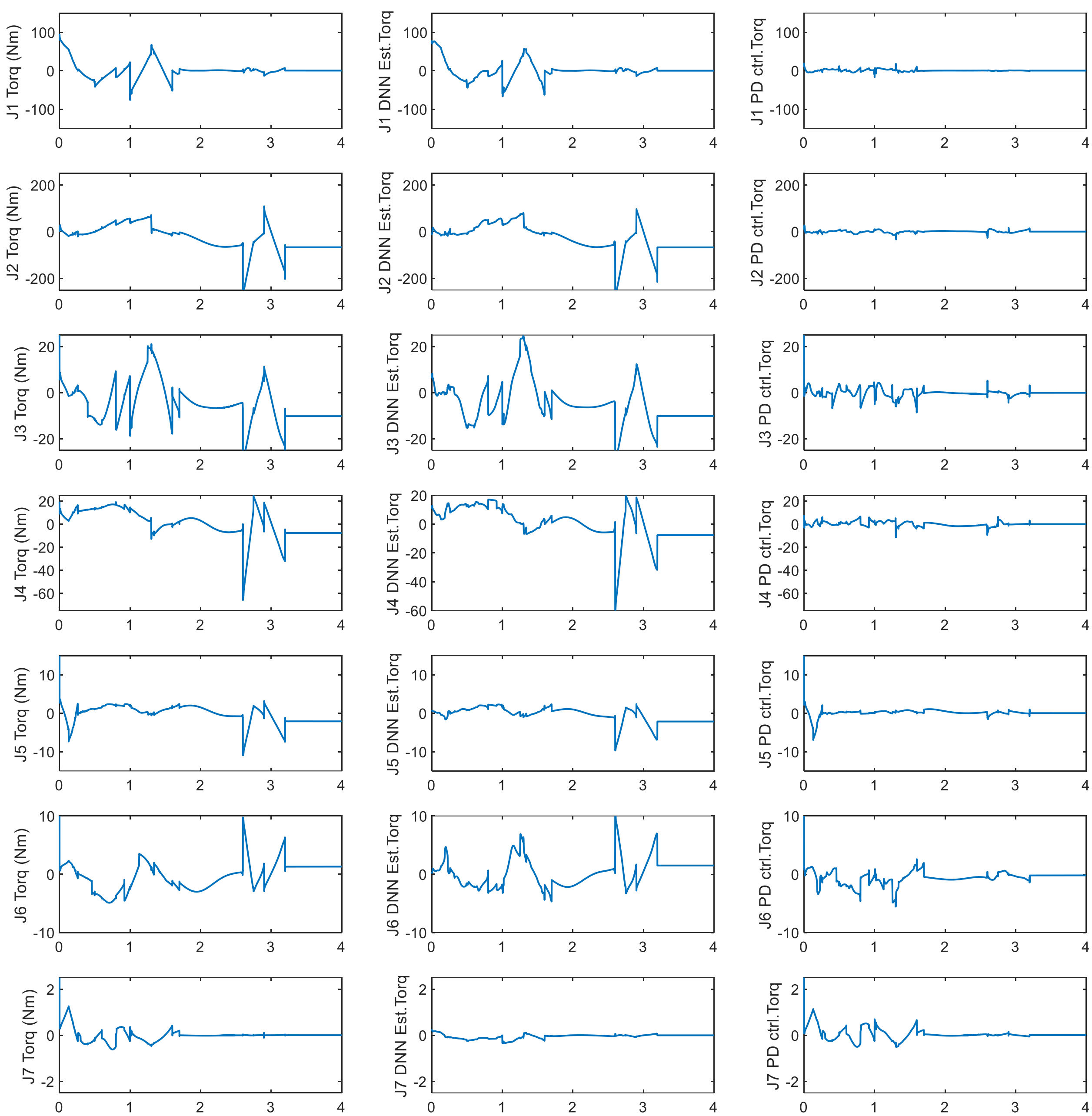

Figure 16 From the right: total Joint torque requirement for sequential joint movements, predicted torque, PD controller's contribution

Figure 16 demonstrates that the torque predicted by the deep neural network closely matches the total required joint torque for all joints, with minimal corrective contribution from the PD controller, except for joint 7. For the seventh joint, the PD controller contributes a slightly higher torque than the feedforward prediction. Owing to the black-box nature of deep neural networks, the internal representation underlying this behavior is difficult to interpret explicitly. It was observed that the torque demand of joint 7 is significantly lower compared to the other joints, which may limit the network's ability to accurately capture its dynamics. Attempts to improve the modeling of joint 7 by adding additional neurons or layers resulted in overfitting, and therefore were not adopted.

The trajectory tracking performance of the developed controller under sequential joint movements is presented in Figure 17. Similar to the simultaneous motion case, the output trajectories closely follow the reference trajectories, making them nearly indistinguishable in the figure. The corresponding trajectory tracking errors are shown in Figure 18. The maximum observed tracking errors for joints 1 through 7 are $[-0.22°, -0.26°, -0.18°, 0.25°, -0.15°, -0.19°, -0.17°]$, respectively. Among these, joint 2 exhibits the largest tracking error of $0.26°$, followed by joints 4 and 1, while all joints maintain low error magnitudes, further validating the effectiveness of the proposed hybrid control strategy.

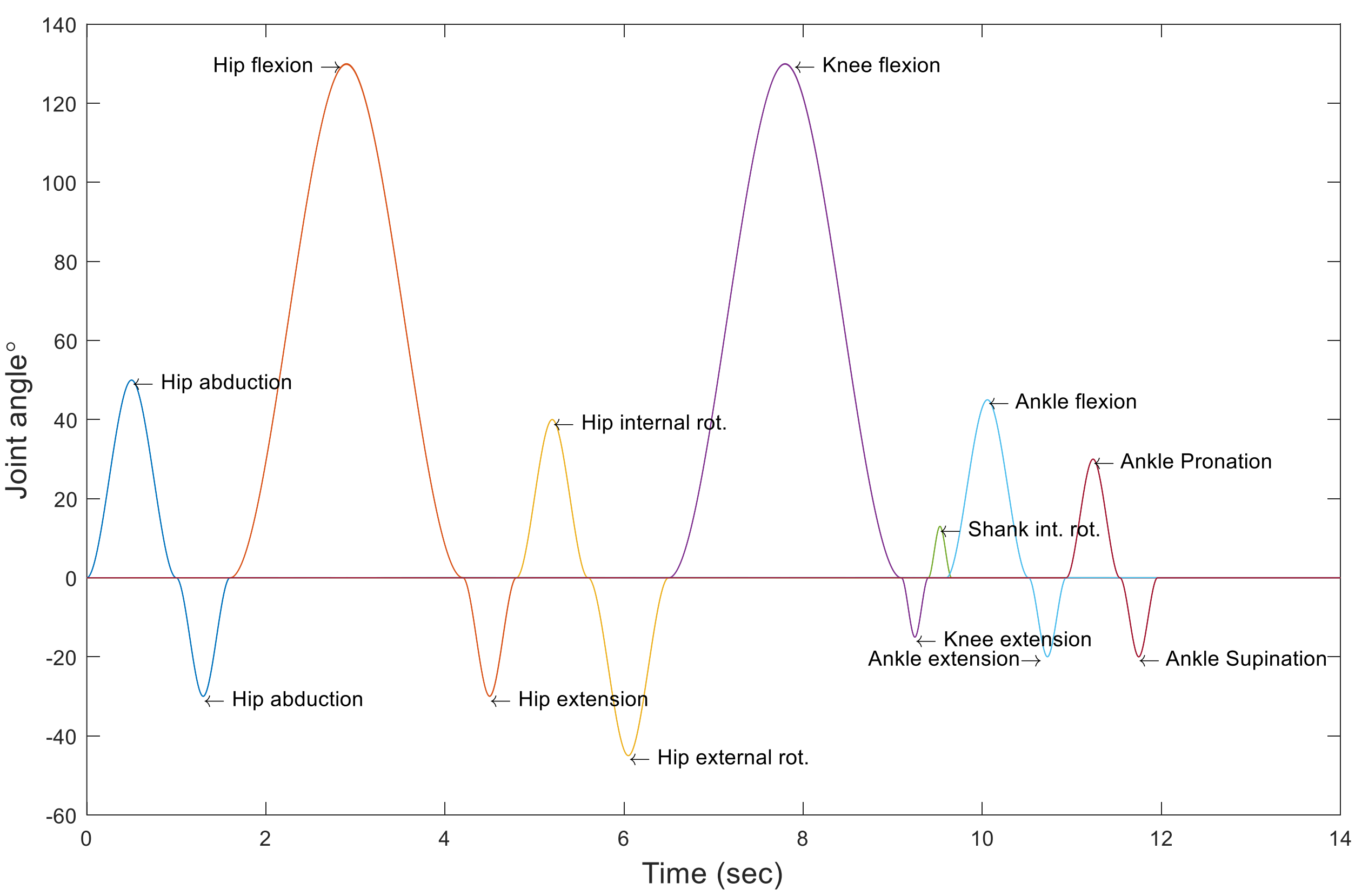

Figure 17 Trajectory tracking performance (Sequential joint movements)

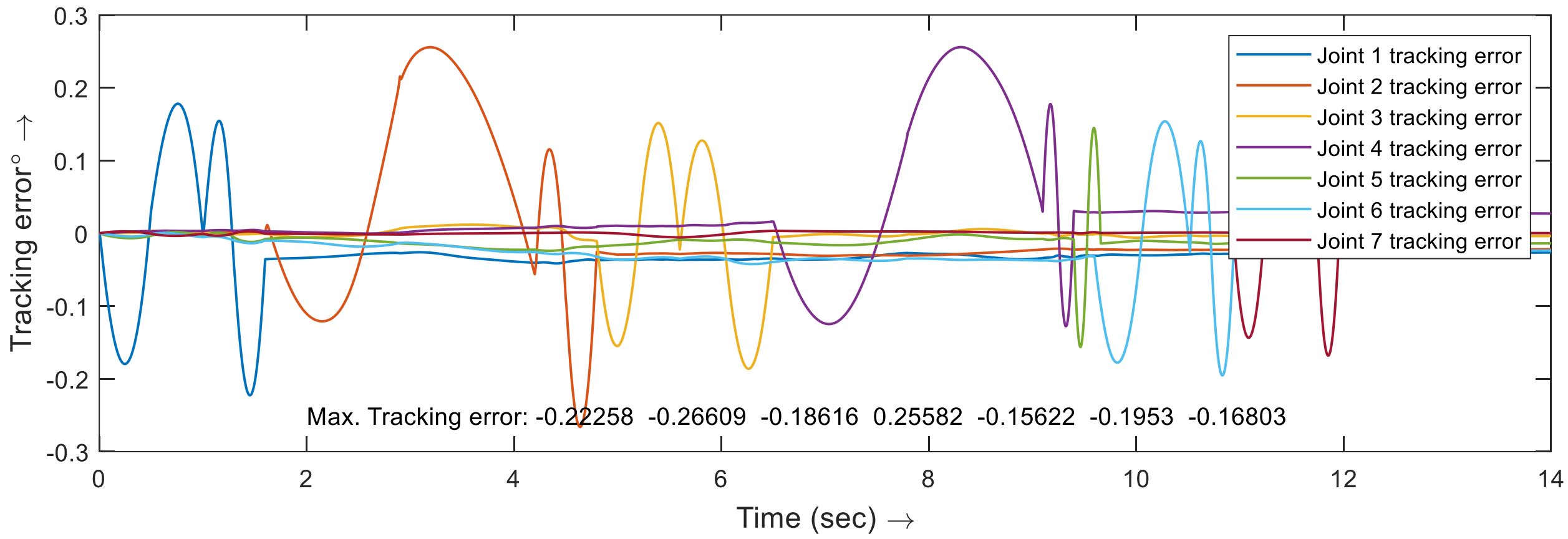

Figure 18 Trajectory tracking error (Sequential joint movements)

Figure 19 presents the joint torque distribution for sequential trajectory tracking. The left column illustrates the total joint torques required to follow the sequential movements, while the middle column shows the torques predicted by the deep neural network. The right column depicts the corrective torque contributions provided by the PD controller. Consistent with the results observed for simultaneous joint movements, the largest prediction error occurs at joint 7. For joints 1 through 6, the torques predicted by the deep neural network closely match the total torque requirements, indicating effective feedforward compensation and minimal reliance on feedback correction.

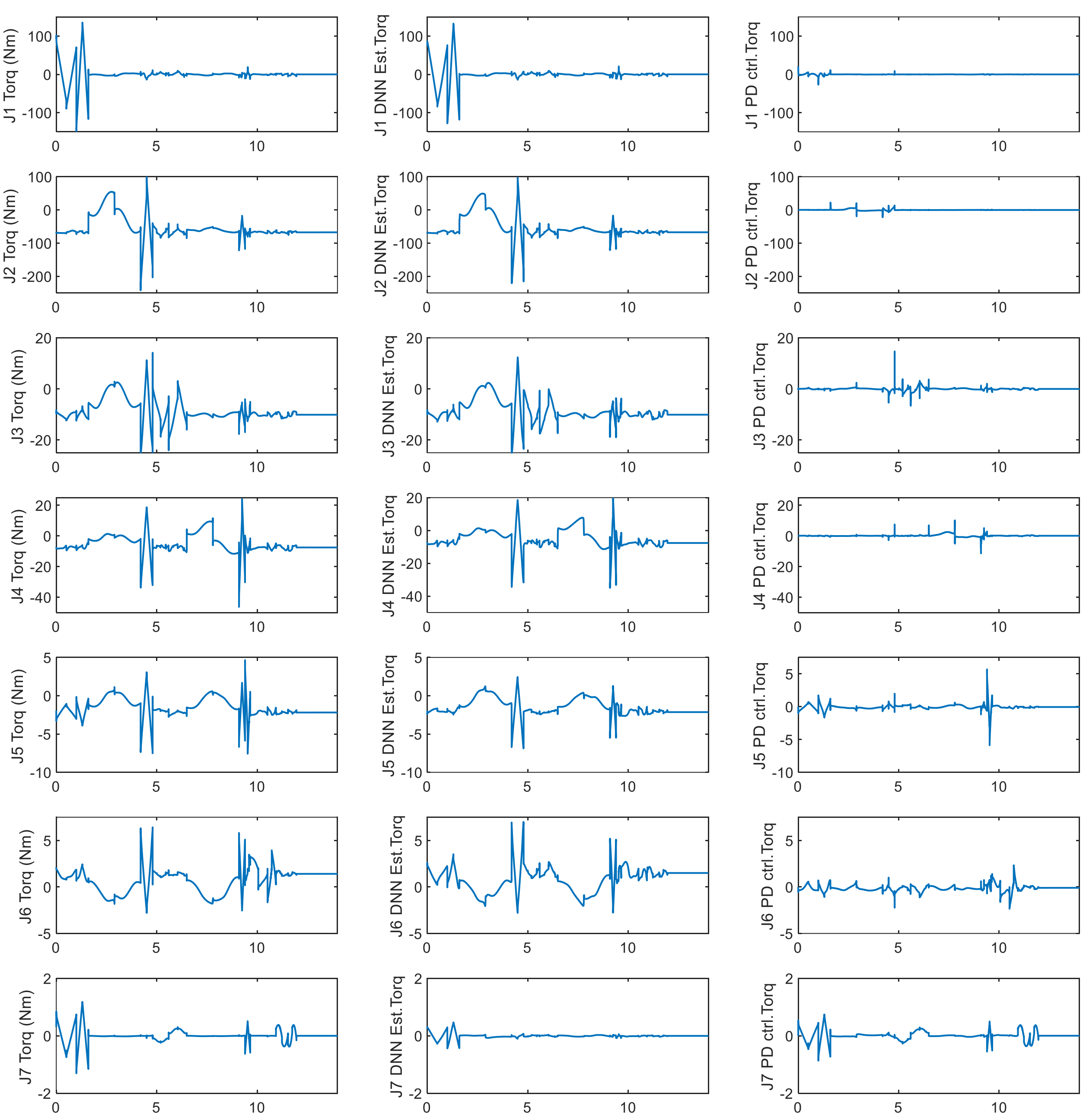

Figure 19 From the left: total Joint torque requirement for sequential joint movements, DNN Predicted torque, PD controller's contribution

Based on the simulation results, the proposed deep neural network–based hybrid controller achieves excellent trajectory tracking performance. The robustness of the controller under parameter variations is examined in the following section.

## 5. Controller robustness analysis against parameter variations ANOVA

Analysis of variance (ANOVA), originally introduced by Ronald Fisher [49], is a statistical method used to evaluate the effect of parameter variations on system outputs. In this study, the user's body mass and height were varied to examine their influence on trajectory tracking errors and to assess the robustness of the proposed controller. A key metric in ANOVA is the *p*-value, which quantifies the strength of evidence against the null hypothesis. When the *p*-value is lower than the selected significance level (0.005), it indicates that the variation of at least one parameter has a statistically significant effect on the output.

A one-way ANOVA test was employed to investigate the relationship between trajectory tracking errors and variations in user mass and height. This test requires that the data be approximately normally distributed and that a sufficiently large sample size be available. Figure 20 and Figure 21 illustrate the distribution of tracking errors corresponding to changes in user weight and height, respectively, and confirm that the error distributions satisfy the normality assumption. The resulting *p*-values for all seven joints are significantly greater than the chosen significance level of 0.005, as summarized in Table 4 and Table 5. These results indicate that variations in user mass and height do not have a statistically significant impact on the controller's trajectory tracking performance, thereby demonstrating the robustness of the proposed control strategy.

The complete form of the variables used in Table 4 and Table 5 are as follows,

***SS*** Sum of squares

***df*** - Degrees of freedom

***MS*** - Mean square

***F*** - statistic

***Prob*** - p-value

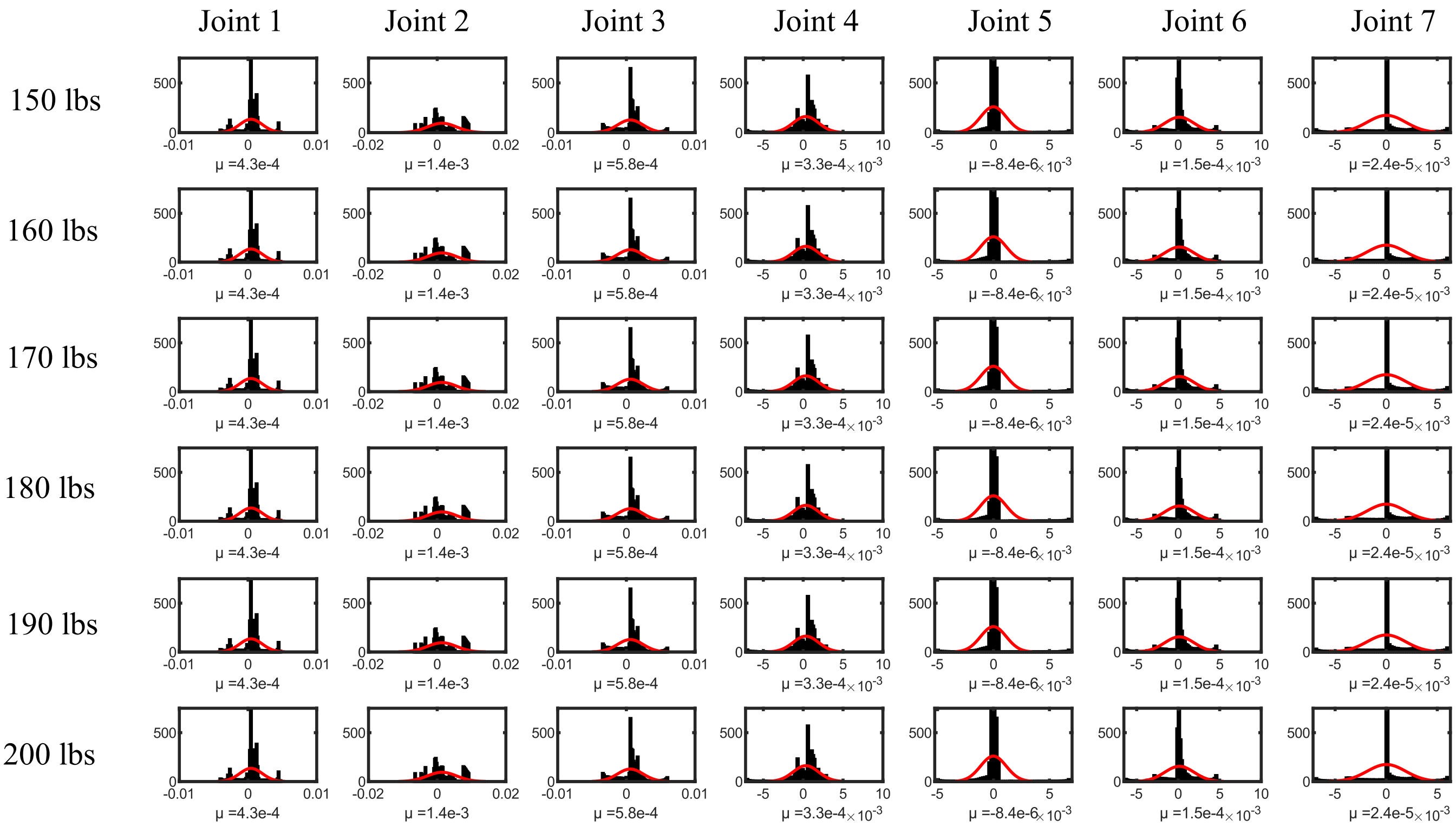

Figure 20 Tracking error distribution concerning weight variation

Table 4 Tracking error variation concerning the user's weight

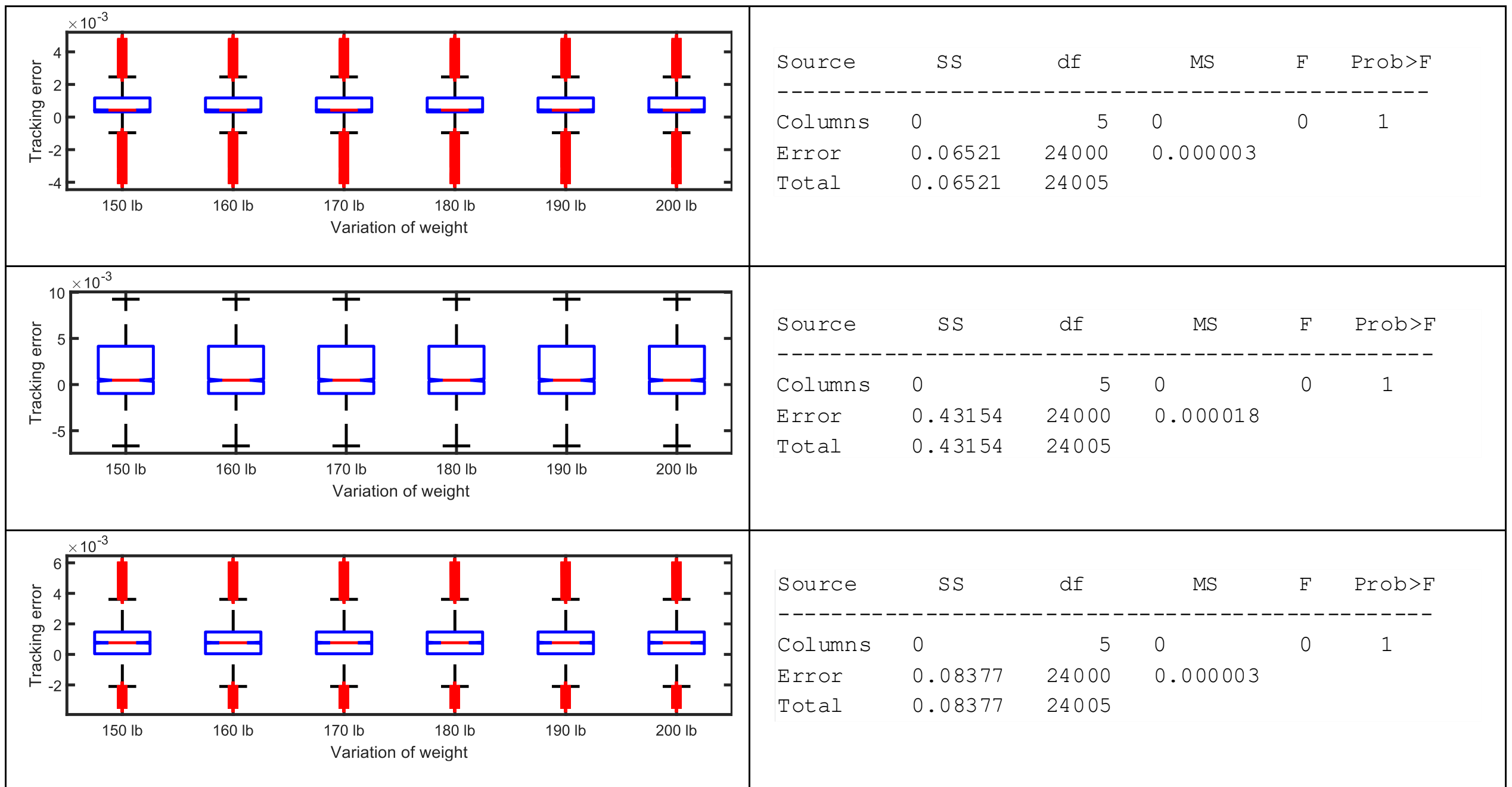

| Source | SS | df | MS | F | Prob>F |
|---|---|---|---|---|---|
| Columns | 0 | 5 | 0 | 0 | 1 |
| Error | 0.06521 | 24000 | 0.000003 | | |
| Total | 0.06521 | 24005 | | | |

| Source | SS | df | MS | F | Prob>F |
|---|---|---|---|---|---|
| Columns | 0 | 5 | 0 | 0 | 1 |
| Error | 0.43154 | 24000 | 0.000018 | | |
| Total | 0.43154 | 24005 | | | |

| Source | SS | df | MS | F | Prob>F |
|---|---|---|---|---|---|
| Columns | 0 | 5 | 0 | 0 | 1 |
| Error | 0.08377 | 24000 | 0.000003 | | |
| Total | 0.08377 | 24005 | | | |

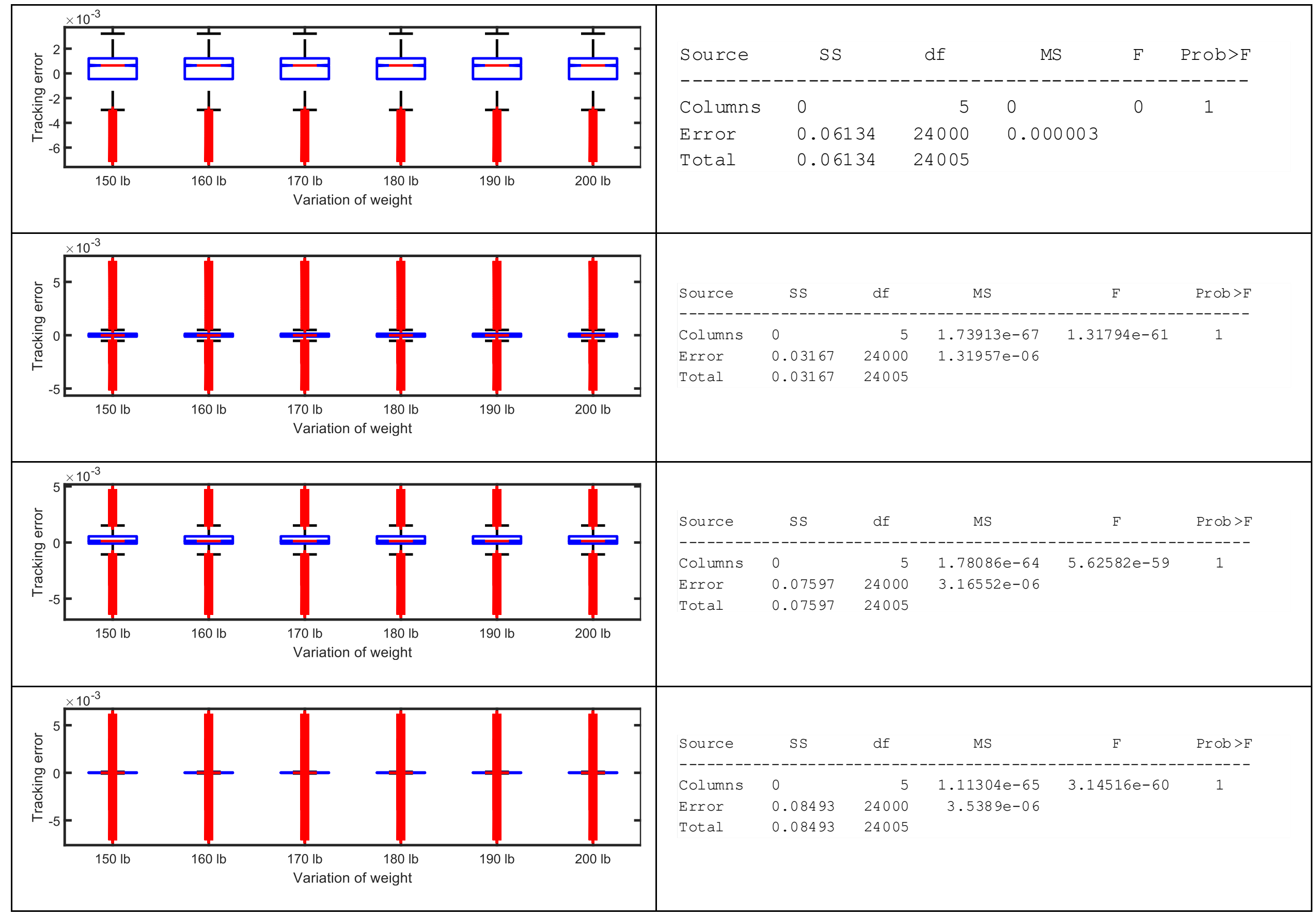

| Source | SS | df | MS | F | Prob>F |
|---|---|---|---|---|---|
| Columns | 0 | 5 | 0 | 0 | 1 |
| Error | 0.06134 | 24000 | 0.000003 | | |
| Total | 0.06134 | 24005 | | | |

| Source | SS | df | MS | F | Prob>F |
|---|---|---|---|---|---|
| Columns | 0 | 5 | 1.73913e-67 | 1.31794e-61 | 1 |
| Error | 0.03167 | 24000 | 1.31957e-06 | | |
| Total | 0.03167 | 24005 | | | |

| Source | SS | df | MS | F | Prob>F |
|---|---|---|---|---|---|
| Columns | 0 | 5 | 1.78086e-64 | 5.62582e-59 | 1 |
| Error | 0.07597 | 24000 | 3.16552e-06 | | |
| Total | 0.07597 | 24005 | | | |

| Source | SS | df | MS | F | Prob>F |
|---|---|---|---|---|---|
| Columns | 0 | 5 | 1.11304e-65 | 3.14516e-60 | 1 |
| Error | 0.08493 | 24000 | 3.5389e-06 | | |
| Total | 0.08493 | 24005 | | | |

Variation of tracking errors with change of subject's height

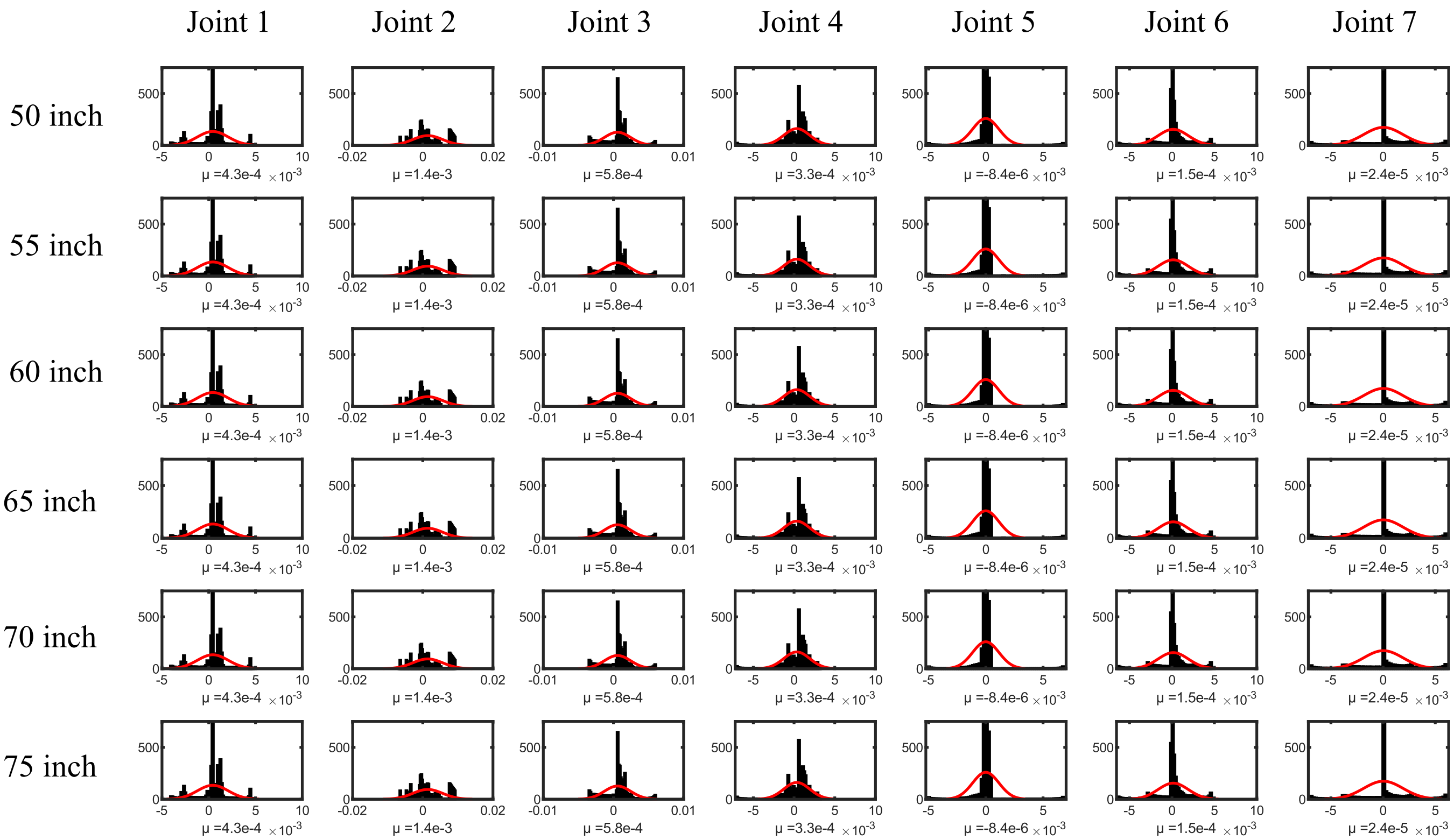

Figure 21 Tracking error distribution concerning the user's height variation

Table 5 Tracking error variation concerning the user's height

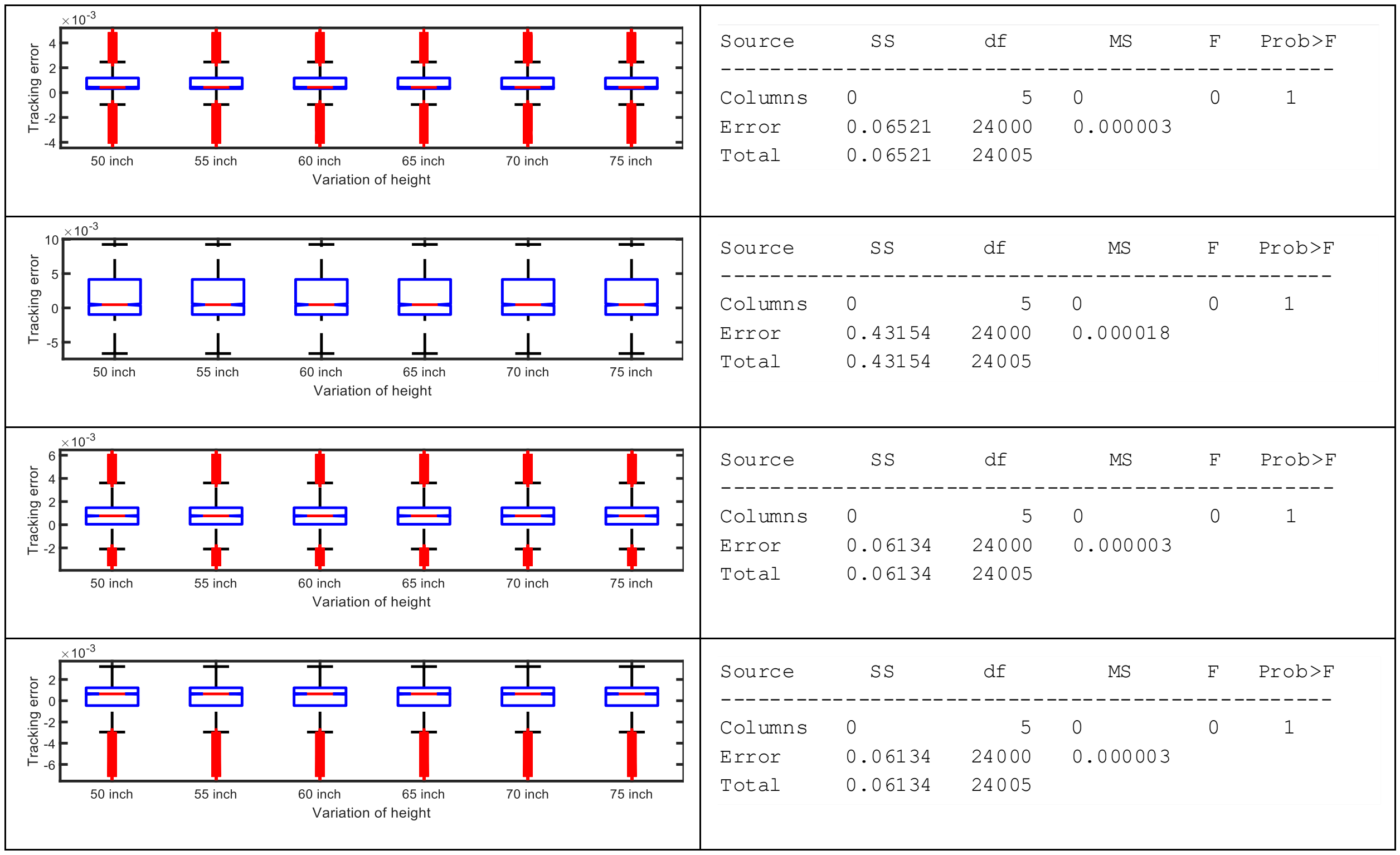

| Source | SS | df | MS | F | Prob>F |
| --- | --- | --- | --- | --- | --- |
| Columns | 0 | 5 | 0 | 0 | 1 |
| Error | 0.06521 | 24000 | 0.000003 | | |
| Total | 0.06521 | 24005 | | | |

| Source | SS | df | MS | F | Prob>F |
| --- | --- | --- | --- | --- | --- |
| Columns | 0 | 5 | 0 | 0 | 1 |
| Error | 0.43154 | 24000 | 0.000018 | | |
| Total | 0.43154 | 24005 | | | |

| Source | SS | df | MS | F | Prob>F |
| --- | --- | --- | --- | --- | --- |
| Columns | 0 | 5 | 0 | 0 | 1 |
| Error | 0.06134 | 24000 | 0.000003 | | |
| Total | 0.06134 | 24005 | | | |

| Source | SS | df | MS | F | Prob>F |
| --- | --- | --- | --- | --- | --- |
| Columns | 0 | 5 | 0 | 0 | 1 |
| Error | 0.06134 | 24000 | 0.000003 | | |
| Total | 0.06134 | 24005 | | | |

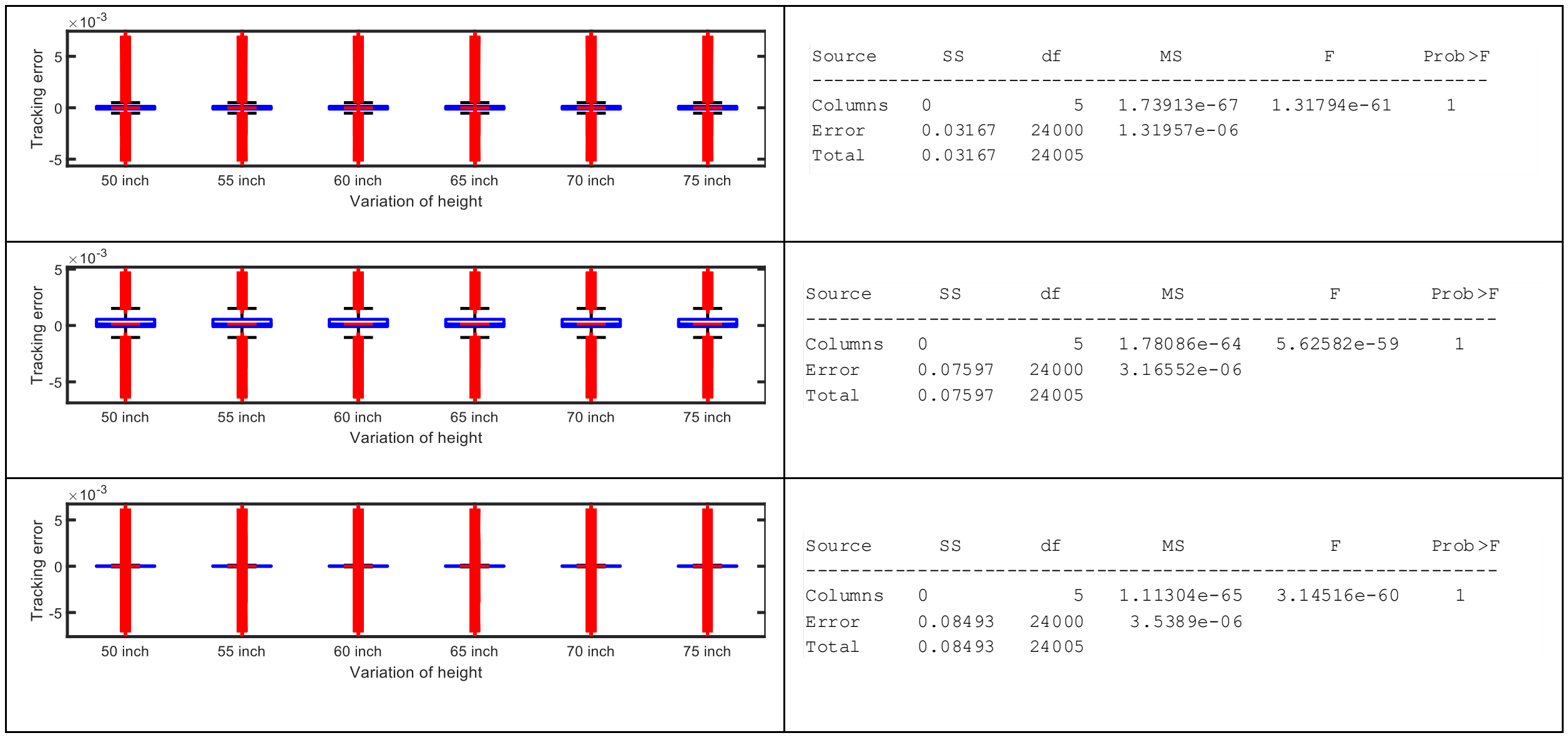

| Source | SS | df | MS | F | Prob>F |
|---|---|---|---|---|---|
| Columns | 0 | 5 | 1.73913e-67 | 1.31794e-61 | 1 |
| Error | 0.03167 | 24000 | 1.31957e-06 | | |
| Total | 0.03167 | 24005 | | | |

| Source | SS | df | MS | F | Prob>F |
|---|---|---|---|---|---|
| Columns | 0 | 5 | 1.78086e-64 | 5.62582e-59 | 1 |
| Error | 0.07597 | 24000 | 3.16552e-06 | | |
| Total | 0.07597 | 24005 | | | |

| Source | SS | df | MS | F | Prob>F |
|---|---|---|---|---|---|
| Columns | 0 | 5 | 1.11304e-65 | 3.14516e-60 | 1 |
| Error | 0.08493 | 24000 | 3.5389e-06 | | |
| Total | 0.08493 | 24005 | | | |

All the ANOVA results summarize on Table 4, Table 5 concludes that the developed controller is highly robust against the syetm parameter variations.

The following section presents a comparative study between the proposed controller and several established control strategies, including the Sliding Mode Controller, Computed Torque Controller, Adaptive Controller, Linear Quadratic Regulator, and Model Reference Computed Torque Controller.

## 6. Performance comparison of the developed Deep Neural Network based controller with Computed Torque Controller, Model Reference Computed Torque Controller , Sliding Mode Controller, Adaptive controller and Linear Quadratic Regulator

Control systems play a critical role in robotic applications by ensuring accurate trajectory tracking and robust performance under uncertainties. This section presents a comparative study between the proposed deep neural network-based hybrid controller and several established robot control strategies, including the Computed Torque Controller (CTC) [8], Model Reference Computed Torque Controller (MRCTC) [13], Sliding Mode Controller (SMC) [14], Adaptive Controller [12], and Linear Quadratic Regulator (LQR) [50], all evaluated using the same robot dynamic model. The objective of this comparison is to assess the effectiveness and practical applicability of the proposed control approach.

Figure 22 illustrates the overall trajectory tracking performance of the developed deep neural network-based controller alongside the CTC, MRCTC, SMC, Adaptive Controller, and LQR. All controllers exhibit high trajectory tracking accuracy, with tracking errors confined within 1°for both sequential and simultaneous joint movements. Due to the small magnitude of the tracking errors, the output trajectories are nearly indistinguishable from the reference trajectories in the figure. A quantitative comparison of the maximum trajectory tracking errors for sequential and simultaneous joint motions across all controllers is provided in Table 6 and Table 7, respectively.

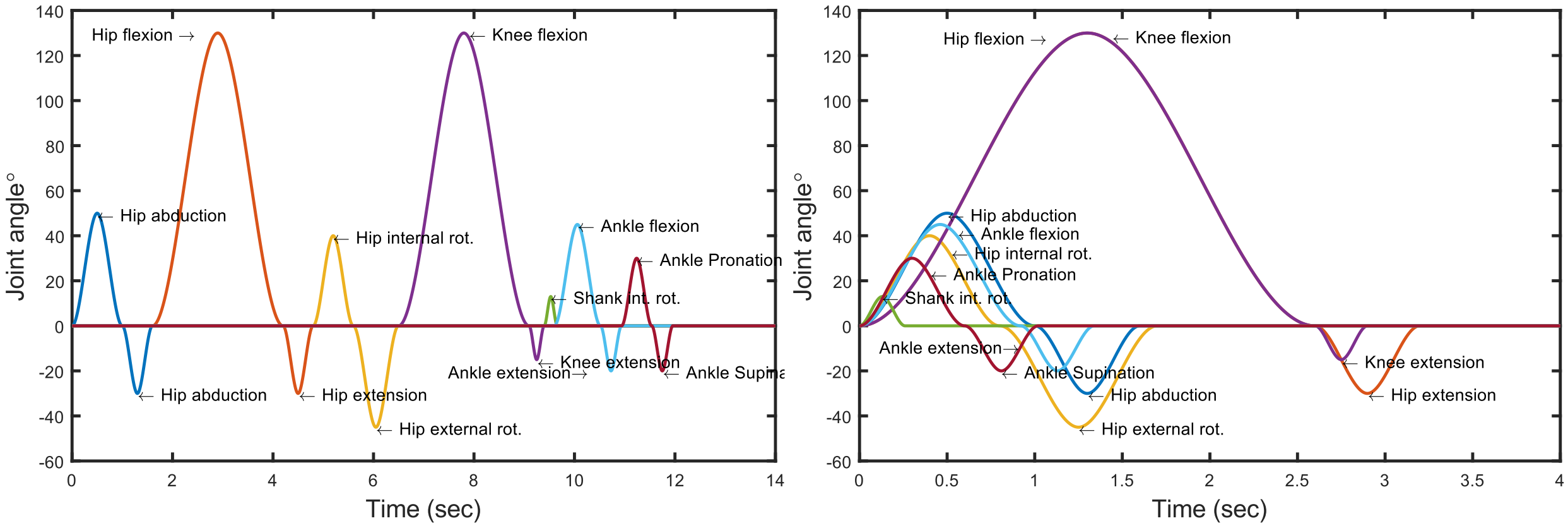

Figure 22 Trajectory tracking performance for the sequential joint movements (left), simultaneous joint movements (right)

Table 6 Maximum values of the trajectory tracking error (Sequential joint movement)

| Control technique | Maximum values of the trajectory tracking error (Sequential joint movement) | | | | | | |
|---|---|---|---|---|---|---|---|
| | Joint 1 | Joint 2 | Joint 3 | Joint 4 | Joint 5 | Joint 6 | Joint 7 |
| Hybrid DNN-PD controller | 0.0558° | .01642° | 0.0304° | 0.1115° | 0.0240° | 0.0501° | 0.0344° |
| CTC | 0.1175° | 0.0909° | 0.0991° | 0.1637° | 0.1694° | 0.1441° | 0.1590° |
| MRCTC | 0.1176° | 0.0910° | 0.0991° | 0.1639° | 0.1694° | 0.1444° | 0.1593° |
| SMC | 0.0396° | 0.0398° | 0.0316° | 0.0614° | 0.0654° | 0.0483° | 0.0482° |
| Adaptive Controller | 0.0727° | 0.0569° | 0.0025° | 0.0136° | 0.0027° | 0.0018° | 0.0005° |
| LQR | 0.3240° | 0.0589° | 0.5050° | 0.1760° | 0.5220° | 0.2400° | 0.1780° |

Table 7 Maximum values of the trajectory tracking error (Simultaneous joint movement)

| Control technique | Maximum values of the trajectory tracking error (Simultaneous joint movement) | | | | | | |
|---|---|---|---|---|---|---|---|
| | Joint 1 | Joint 2 | Joint 3 | Joint 4 | Joint 5 | Joint 6 | Joint 7 |
| Hybrid DNN-PD controller | 0.1181° | 0.0762° | 0.0565° | 0.1132° | 0.0231° | 0.0767° | 0.0369° |
| CTC | 0.1175° | 0.0909° | 0.1041° | 0.1637° | 0.1697° | 0.1467° | 0.1734° |
| MRCTC | 0.1176° | 0.0910° | 0.1041° | 0.1639° | 0.1697° | 0.1470° | 0.1736° |
| SMC | 0.3966° | 0.3988° | 0.3165° | 0.6142° | 0.6545° | 0.4838° | 0.4842° |
| Adaptive Controller | 0.0378° | 0.0762° | 0.0365° | 0.0208° | 0.0726° | 0.0022° | 0.0010° |
| LQR | 0.6450° | 0.1060° | 0.8000° | 0.2800° | 0.6180° | 0.3010° | 0.8640° |

All six controllers are used to simulate the same robot dynamics. By comparing the controllers, it has been noticed that all the controller's trajectory tracking accuracy is very high. Developed Deep neural network based controller is highly efficient.

Figure 23 to Figure 28 illustrate the joint torque requirements for tracking both sequential and simultaneous joint movements using different control strategies. Figure 23 presents the joint torque profiles obtained with the proposed deep neural network-based hybrid controller. Figure 24 shows the corresponding torque requirements using the Computed Torque Controller. The torque profiles generated by the Model Reference Computed Torque Controller, Sliding Mode Controller, Adaptive Controller, and Linear Quadratic Regulator are presented in Figure 25, Figure 26, Figure 27, and Figure 28 respectively.

Across all control approaches, the peak joint torques are comparable, and the resulting torque profiles exhibit similar trends for both sequential and simultaneous joint movements. These results indicate that the proposed deep neural network-based hybrid controller achieves torque requirements consistent with established control strategies while maintaining comparable actuation demands.

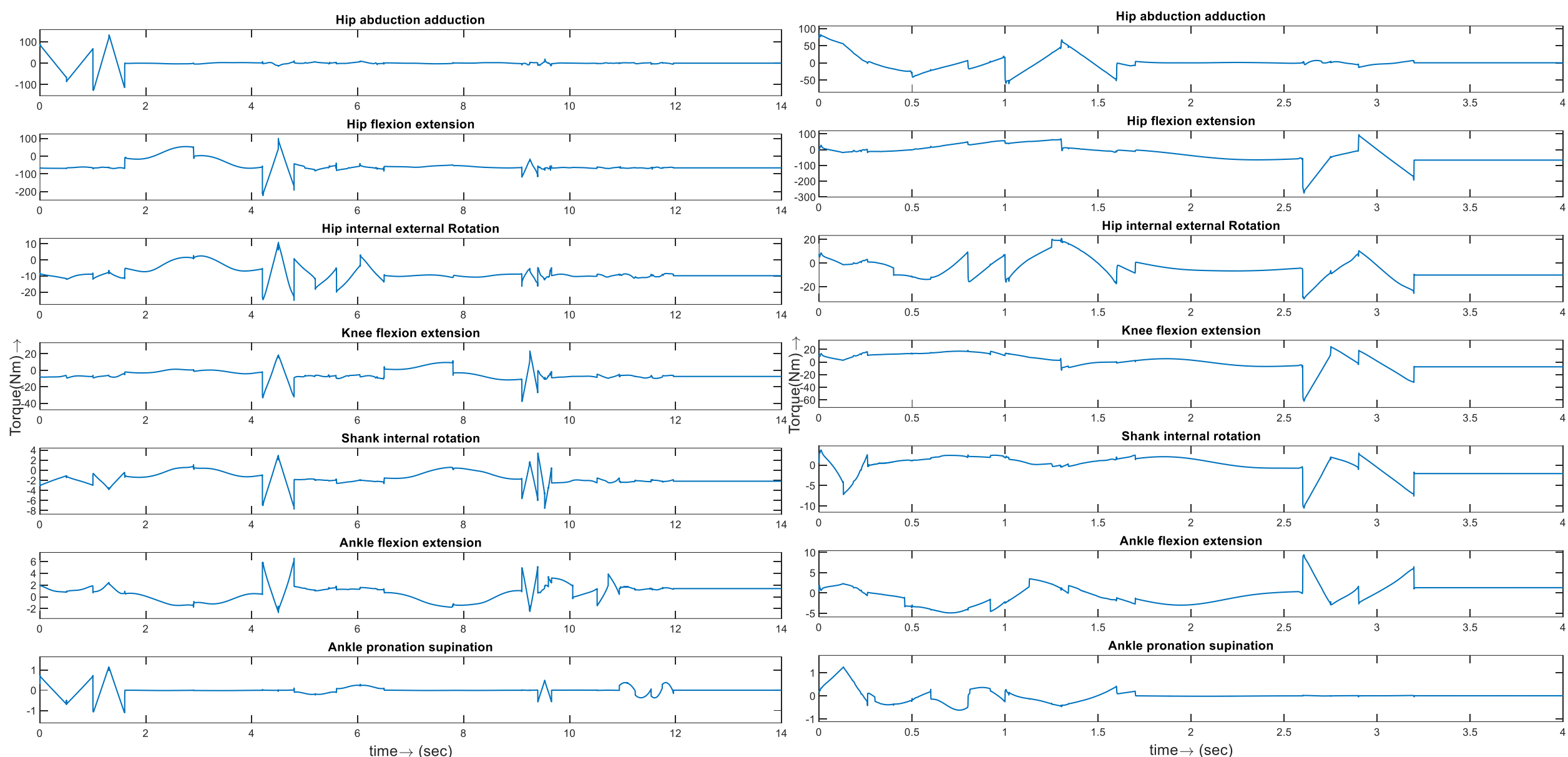

Figure 23 Joint torque requirement for Sequential and Simultaneous Joint Movement using Deep neural network based hybrid controller

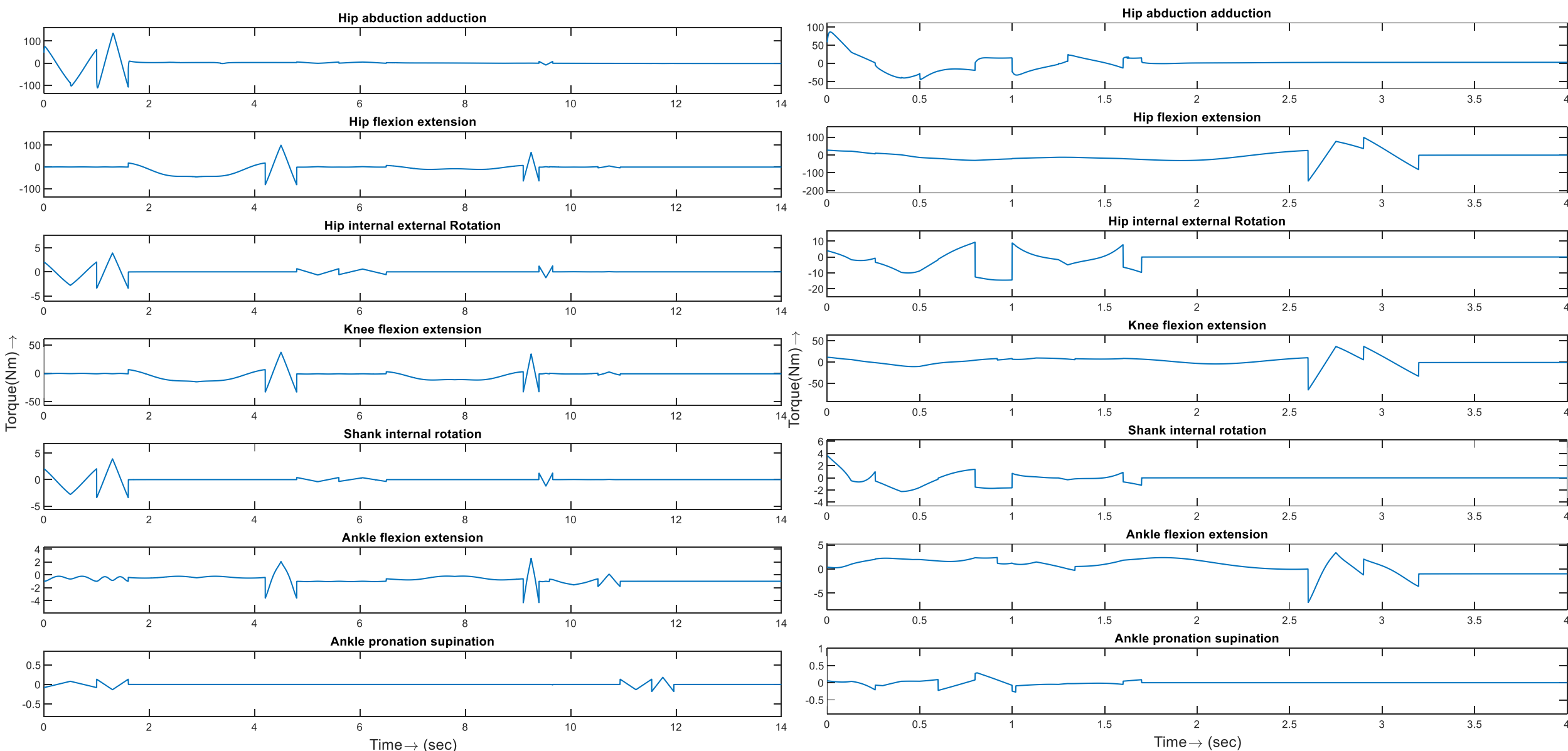

Figure 24 Joint torque requirement for Sequential and Simultaneous Joint Movement using Computed torque controller

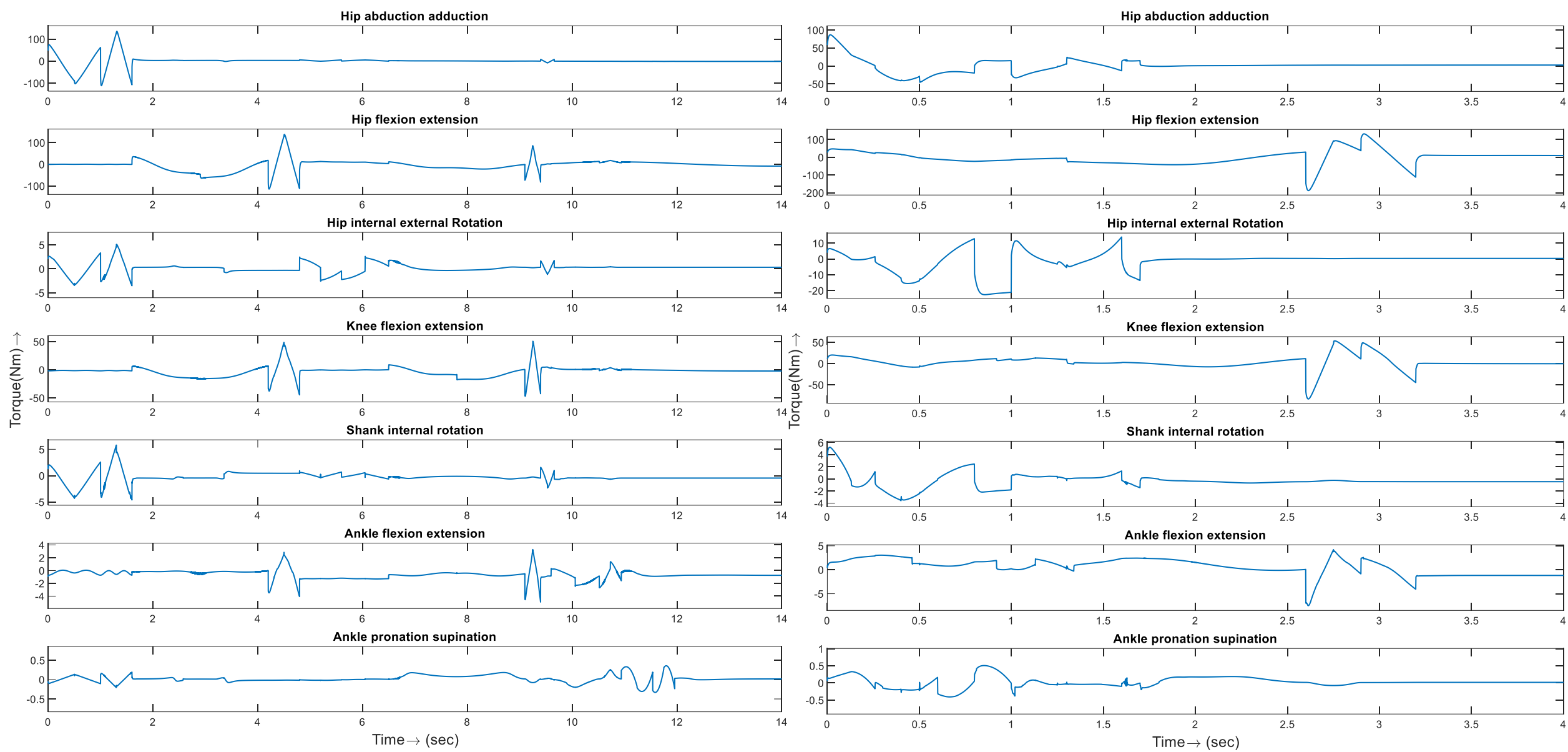

Figure 25 Joint torque requirement for Sequential and Simultaneous Joint Movement using Model reference computed torque controller

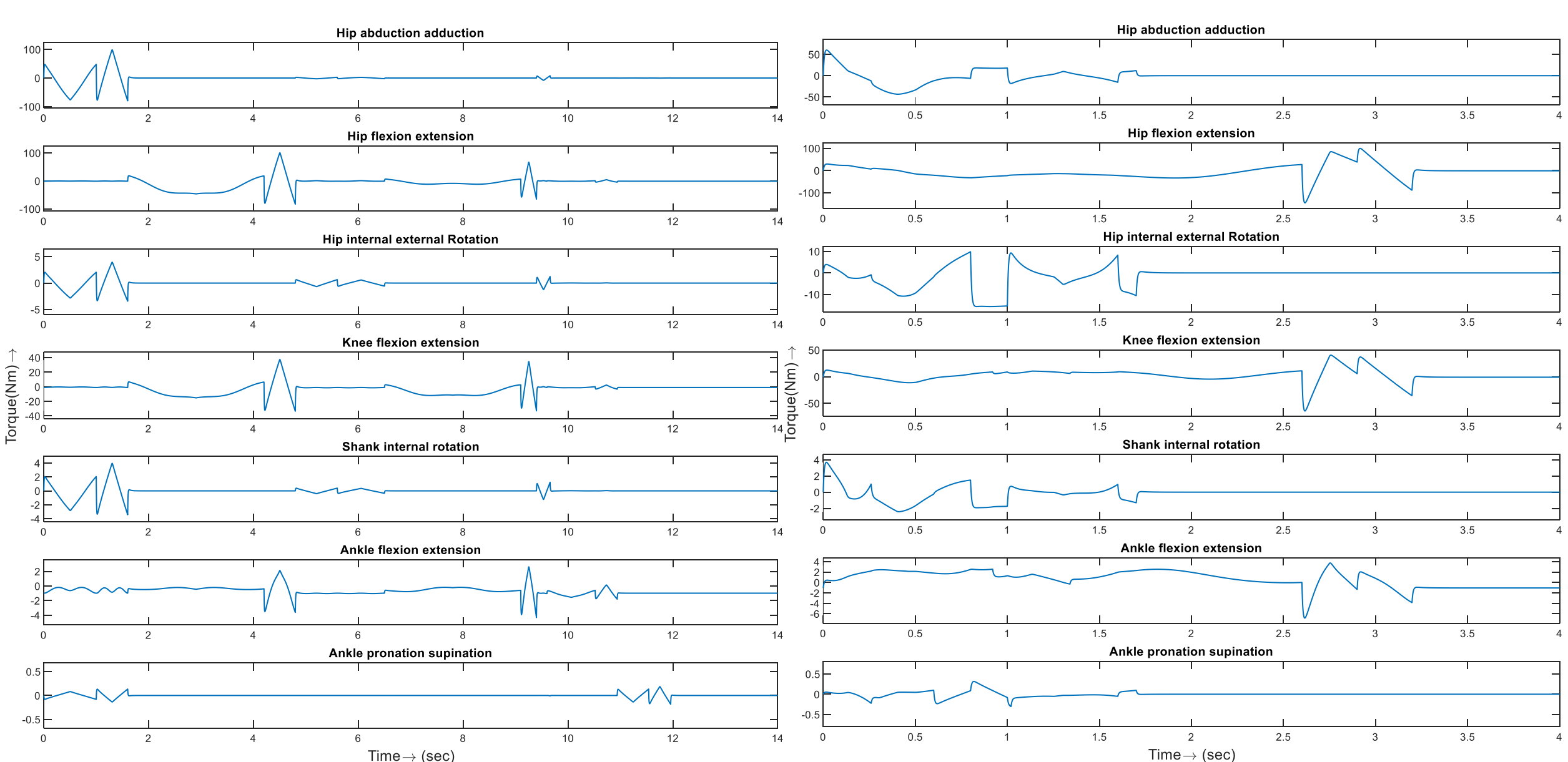

Figure 26 Joint torque requirement for Sequential and Simultaneous Joint Movement using Sliding mode controller

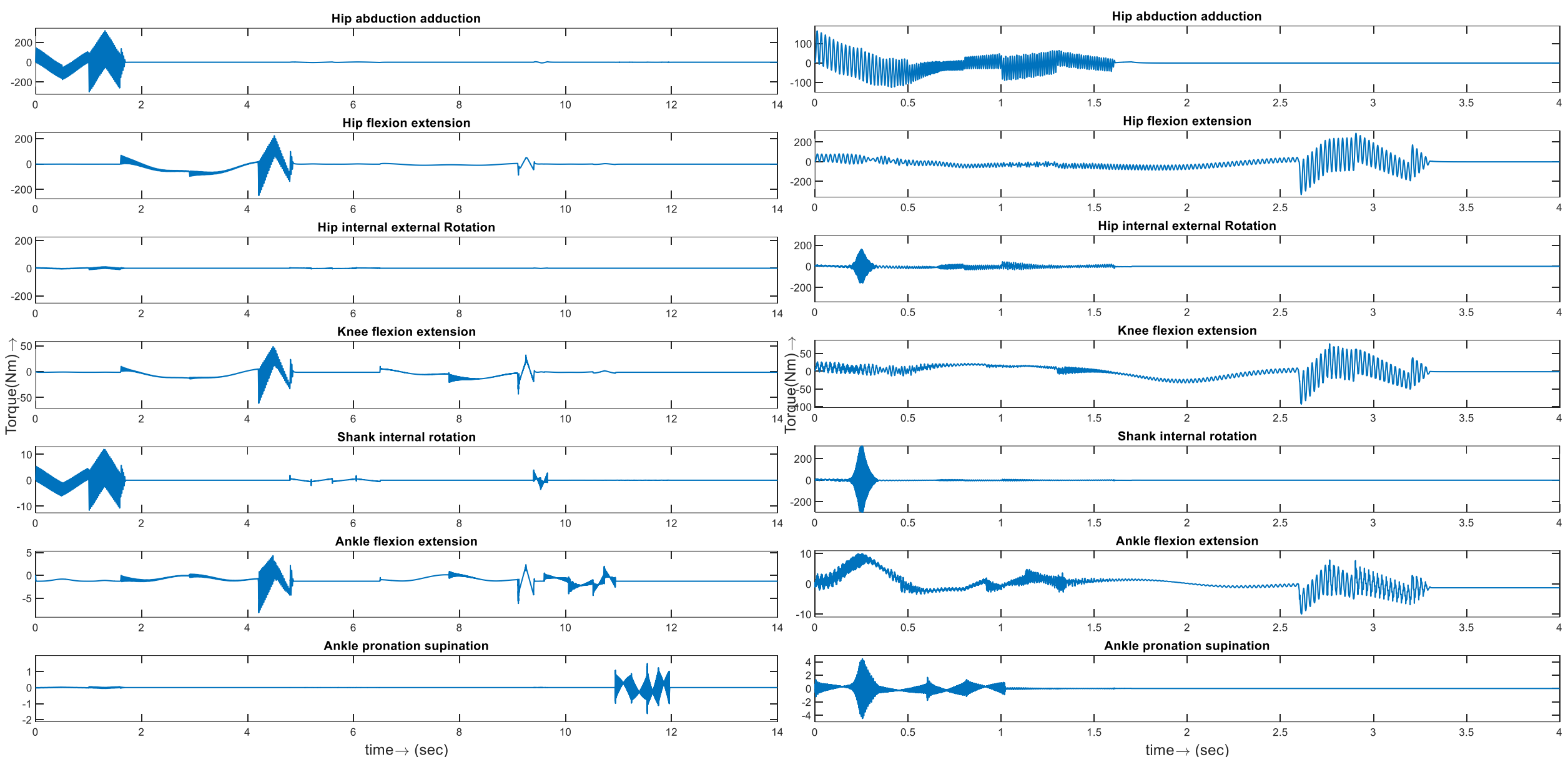

Figure 27 Joint torque requirement for Sequential and Simultaneous Joint Movement using Adaptive controller

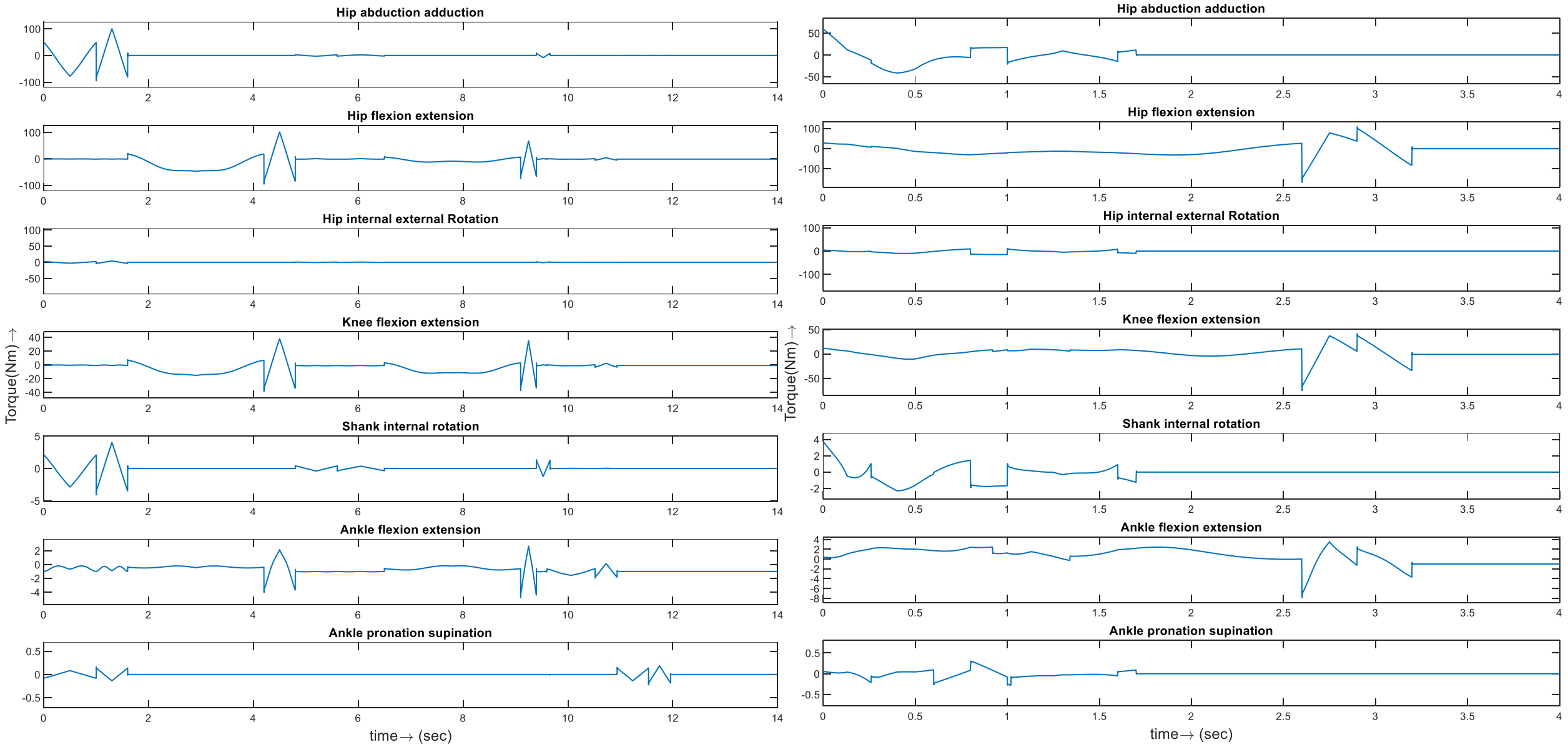

Figure 28 Joint torque requirement for Sequential and Simultaneous Joint Movement using Linear quadratic regulator

Finally, by examiningTable 6, Table 7 and Figure 23 to Figure 28 it can be concluded that the developed deep neural network based controller offers very high trajectory tracking accuracy while using torques similar to the conventional controllers.

## 7. Conclusion

Deep neural networks have emerged as effective tools for dynamic modeling and control of complex robotic systems due to their ability to approximate highly nonlinear dynamics while maintaining low computational cost during real-time execution. In this study, a dynamic model of a human lower extremity exoskeleton robot was developed and controlled using a computed torque–based framework. The dynamic system was simulated under a wide range of reference trajectories and subject-specific parameters, including user height and weight. The resulting joint torque data, together with the reference trajectories and user parameters, were used to train a deep neural network.

A four-layer deep neural network architecture was designed to predict joint torque requirements based on reference trajectories and user anthropometric information. The use of multiple hidden layers enabled the network to capture abstract features of the highly nonlinear multi-input multi-output dynamics of the exoskeleton system. The trained deep neural network was integrated into a hybrid control framework as a feedforward controller, while a proportional derivative controller was employed in the feedback loop to compensate for residual prediction errors.

Simulation results demonstrated that the proposed deep neural network based hybrid controller achieves excellent trajectory tracking performance for both sequential and simultaneous joint movements. The stability of the closed-loop system was analytically verified, and robustness to variations in user mass and height was statistically validated using ANOVA. Furthermore, comparative studies against several established control strategies, including the Computed Torque Controller, Model Reference Computed Torque Controller, Sliding Mode Controller, Adaptive Controller, and Linear Quadratic Regulator, confirmed that the proposed approach delivers comparable or superior tracking accuracy while requiring similar joint torque levels.

Accurate modeling of complex dynamic systems remains inherently challenging, particularly in the presence of time-varying parameters and unmodeled uncertainties. While the proposed approach effectively addresses these challenges through data-driven feedforward compensation, future work will focus on enhancing adaptability by incorporating reinforcement learning techniques to tune neural network parameters online. This extension has the potential to further improve controller performance and robustness in real-world deployment scenarios.

## Disclosure of interest

The authors report no conflicts of interest. The authors alone are responsible for the content and writing of this article.

## Funding

No funding was received for this research.